\pdfoutput=1

\documentclass[11pt]{article}

\usepackage{acl}

\usepackage{times}
\usepackage{latexsym}

\usepackage[T1]{fontenc}

\usepackage[utf8]{inputenc}

\usepackage{microtype}

\usepackage{amsfonts,bm}

\usepackage{bbm}
\usepackage{nccmath}
\usepackage{comment}
\usepackage{multirow}
\usepackage{caption}
\usepackage{subcaption}
\usepackage{booktabs}
\usepackage[noabbrev,capitalize,nameinlink]{cleveref}

\usepackage{arydshln}
\usepackage{tcolorbox}
\tcbuselibrary{breakable,xparse,skins}

\usepackage{wrapfig}

\usepackage{tikz}
\usepackage{tikz-dependency}

\newcommand{\peach}{\raisebox{-0.03cm}{\includegraphics[width=0.35cm]{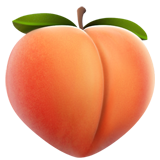}}}

\usepackage[textsize=tiny]{todonotes}

\usepackage{comment}

\usepackage{xcolor,pifont}

\newcommand{\gcheck}{{\color{green!55!black}\ding{52}}}
\newcommand{\rcross}{{\color{red!70!black}\ding{55}}}
\definecolor{textbgcolor}{HTML}{F5F5F5}
\definecolor{promptbgcolor}{HTML}{ecf9ec}
\definecolor{slotcolor}{HTML}{b30000}

\title{Calibrating Large Language Models Using Their Generations Only}

\author{Dennis Ulmer\textsuperscript{1, 2, 3}  \hspace{.5em} Martin Gubri\textsuperscript{1} \hspace{.5em} Hwaran Lee\textsuperscript{4}  \hspace{.5em} Sangdoo Yun\textsuperscript{4}  \hspace{.5em} Seong Joon Oh\textsuperscript{1, 5, 6}\\
\textsuperscript{1}Parameter Lab \textsuperscript{2}IT University of Copenhagen \textsuperscript{3}Pioneer Centre for Artificial Intelligence\\ \textsuperscript{4}NAVER AI Lab \textsuperscript{5}University of T\"{u}bingen \textsuperscript{6}T\"ubingen AI Center\\
\texttt{dennis.ulmer@mailbox.org}
}

\begin{document}
\maketitle
\begin{abstract}
    As large language models (LLMs) are increasingly deployed in user-facing applications, building trust and maintaining safety by accurately quantifying a model's confidence in its prediction becomes even more important.
    However, finding effective ways to calibrate LLMs---especially when the only interface to the models is their generated text---remains a challenge.
    We propose APRICOT \peach\ (\underline{A}uxiliary \underline{pr}ed\underline{i}ction of \underline{co}nfidence \underline{t}argets): A method to set confidence targets and train an additional model that predicts an LLM's confidence based on its textual input and output alone.
    This approach has several advantages: It is conceptually simple, does not require access to the target model beyond its output, does not interfere with the language generation, and has a multitude of potential usages, for instance by verbalizing the predicted confidence or adjusting the given answer based on the confidence.
    We show how our approach performs competitively in terms of calibration error for white-box and black-box LLMs on closed-book question-answering to detect incorrect LLM answers.
\end{abstract}

\section{Introduction}
When given a case description of ``\emph{A man superglued his face to a piano and he says it's making it hard to get a full night of sleep}'', a recently released medical LLM was found to list unrelated potential causes in its diagnosis, including narcolepsy, sleep apnea and others.\footnote{\url{https://x.com/spiantado/status/1620459270180569090} (last accessed Nov. 7, 2023).}
This, of course, ignores the seemingly obvious reason for the patient's complaints.
While humorous, this example illustrates the pitfalls of practical LLM applications:
Despite often looking convincing on the surface---especially to non-experts---model responses can be wrong or unreliable, leading to potentially harmful outcomes or a loss of trust in the system, foregoing its benefits.
Indeed, consistent behavior (imagine e.g.\@ reliably indicating a lack of confidence for unsure responses) has been argued as one way to build trust in automated systems \citep{jacovi2021formalizing}, while misleading predictions have been empirically shown to lead to a loss of trust that can be hard to recover from \citep{dhuliawala2023diachronic}.

\begin{figure}[tb!]
    \centering
    \includegraphics[width=0.96\columnwidth]{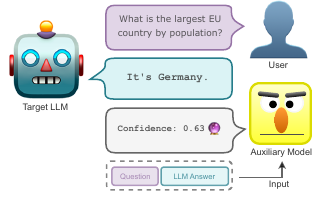}
    \caption{Illustration of APRICOT \peach: We train an auxiliary model to predict a target LLM's confidence based on its input and the generated answer.}
\end{figure}

We introduce APRICOT \peach, a method to use an auxiliary model to infer a LLM's confidence in an  open-ended question-answering setting.
The auxiliary model does so based on the given input to and generated output text from the LLM alone.
The model is trained by using these two parts as input and predicting calibration targets.
The latter are obtained without access to the LLM's sequence likelihoods or internal states by clustering input representations produced by an additional embedding model, and thus only require black-box access.
This especially relevant since an increasing number of LLM providers safeguard their model behind black-box APIs.
This approach is conceptually straightforward, easy to optimize, and opens up a large number of possible applications, for instance, verbalizing uncertainty---a recently popular way to communicate uncertainty for LLMs \citep{lin2022teaching, xiong2023can}---or adjusting a model's response with linguistic markers of confidence.

Our contributions are as follows:
We propose to obtain calibration targets without requiring any additional information about LLM internals or question metadata.
We show that using auxiliary models on the target LLM's input and output is sufficient to predict a useful notion of confidence.
We also perform additional studies to identify which parts of the LLM's output are most useful to predict confidence.
All the code is openly available.\footnote{\url{https://github.com/parameterlab/apricot}.}

\section{Related Work}\label{sec:related-work}

\paragraph{Trustworthiness in ML systems.} 
Trustworthiness has been identified as a key challenge to exploit the potential applications of automated systems \citep{marcus2020next, goldblum2023perspectives}.
This trait is seen as especially under risk in the presence of out-of-distribution examples or distributional shift \citep{ovadia2019can, d2022underspecification}. 
As illustrated by the introductory example, addressing these problems is paramount in high-stakes domains such as healthcare \citep{he2019practical, ulmer2020trust, van2023intensive}, analyzing asylum cases \citep{nalbandian2022eye}, or legal deliberations \citep{chalkidis2023chatgpt, dahl2024large}.
\citet{jacovi2021formalizing} argue that trust in automated systems can for instance be built extrinsically, e.g.\@ through consistent and predictable behavior.
One such a way is \emph{uncertainty quantification}, i.e.\@ by supplying a user with scores reflecting the reliability of a model prediction \citep{bhatt2021uncertainty, liao2022designing, hasan2023survey}.
However, the lab experiments of \citet{dhuliawala2023diachronic} also demonstrate the flip side of this:
When exposing human participants to unreliable confidence estimates, they measure a decrease in trust and participant outcomes alike.
Therefore, a method like ours can help to build trust in LLMs and pave the way for reaping their benefits.

\paragraph{Uncertainty Quantification for LLMs.} 
While methods for predictive uncertainty quantification have already been explored in NLP for classification \citep{ulmer2022exploring, van2022benchmarking, vazhentsev2023hybrid}, regression \citep{beck-etal-2016-exploring, glushkova-etal-2021-uncertainty-aware, zerva2022disentangling} and language generation tasks \citep{xiao2020wat, malinin2021uncertainty}, their application to LLMs has posed novel challenges.
Besides the different kinds of uncertainty due to the nature of language itself \citep{baan2023uncertainty}, LLMs are usually too large for Bayesian methods and are more expensive to be finetuned (unless one resorts to low-rank approximations; \citealp{yang2023bayesian}). 
Furthermore, the generation process is often shielded behind black-box APIs for commercial models, only leaving access to the predicted probabilities, or---in the worst case---the generated text.
Existing approaches operate on the model's confidence and improve it through (re-)calibration \citep{tian2023just, chen2024reconfidencing, bakman2024mars}, the ensembling of prompts \citep{jiang2023calibrating, hou2023decomposing}, asking the model to rate its own uncertainty \citep{lin2022teaching, chen2023quantifying, tian2023just} or analyzing its use of linguistic markers \citep{zhou2023navigating}, computing the entropy over sets of generations with similar meaning \citep{kuhn2023semantic} or comparing the textual similarity of generations for the same input \citep{lin2023generating}.
The most similar work to ours comes from \citet{mielke2022reducing}, who train a calibrator on the target model's hidden states to predict whether the answer is correct or incorrect.
They then finetune the target model using control tokens that are based on the calibrator's prediction and indicate the desired confidence level.
In comparison, our method has multiple advantages: 
We only assume access to the input question and text outputs, not requiring white-box model access or finetuning of the target model.
We further demonstrate in our experiments that using more fine-grained targets than the simple binary yields better calibration overall.

\paragraph{Predicting properties of generated text.} 
Instead of trying to obtain a piece of information of interest---e.g.\@ truthfulness or predictive confidence---from the original model, other works have investigated whether this can be done so through secondary (neural) models. 
For instance, \citet{lahlou2023deup, mukhoti2023deep} fit density estimators on internal model representations.
In NLP, \citet{pacchiardi2023catch} demonstrated that wrong statements by LLMs can be detected by collecting yes / no answers for a set of questions, storing them in a vector and fitting a logistic regression model. 
In general, using secondary neural models to predict properties of the generated text also has connections to other tasks such as translation quality estimation \citep{blatz2004confidence, quirk2004training, wang2019improving, glushkova-etal-2021-uncertainty-aware, zerva2022better}, toxicity classification \citep{maslej2020comparison} or fine-grained reward modeling \citep{wu2023fine}.

\section{Methodology}\label{sec:methodology}

\begin{table}[htb]
    \centering
    \renewcommand{\arraystretch}{1.35}
    \resizebox{.995\columnwidth}{!}{
    \begin{tabular}{@{}rccc@{}}
    \toprule
    Method & Black-box LLM? & Consistent? & Calibrated? \\
    \midrule
    Seq. likelihoods & \rcross & \gcheck & \rcross \\
    Verb. uncertainty & \gcheck & \rcross & \rcross \\
    APRICOT \peach\ (ours) & \gcheck & \gcheck & \gcheck \\
    \bottomrule
    \end{tabular}%
    }
    \caption{Comparison of appealing attributes that LLM confidence quantification techniques should fulfil. They should ideally be applicable to black-box LLMs, be consistent (i.e., always elicit a response), and produce calibrated estimates of confidence.}\label{tab:constrasting-methods}
\end{table}

\noindent Estimating the confidence of a LLM can be challenging, since their size rules out many traditional techniques that require finetuning or access to model parameters.
In this light, using the likelihood of the generated sequence might seem like an appealing alternative;
however, it might not actually reflect the reliability of the model's answer and often cannot be retrieved when using black-box models, where the only output is the generated text.
Verbalized uncertainty, i.e.\@ prompting the LLM to express its uncertainty in words, can be a solution when the model is powerful enough.
But as we later show in \cref{sec:experiments}, the generated confidence expressions are not very diverse, and results are not always \emph{consistent}, meaning that the model does not always generate a desired confidence self-assessment.
As we illustrate in \cref{tab:constrasting-methods}, our method, APRICOT \peach, fulfills all of these criteria:
Through a one-time finetuning procedure of an auxiliary model on the target LLMs outputs, we have full control over a calibrated model that gives consistent and precise confidence estimates.

\begin{figure}[htb]
    \centering
    \includegraphics[width=0.995\columnwidth]{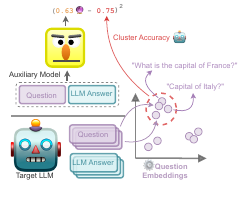}
    \caption{Full overview of APRICOT \peach. We collect a LLM's answer to a set of questions and embed the latter using an embedding model. After clustering similar questions and identifying the LLM's accuracy on them, we can use this value as reference when training to predict the confidence from a question-answer pair.}\label{fig:method}
\end{figure}

In \cref{fig:method} we give an overview of APRICOT \peach, which consists of three main steps:
Firstly, we prompt the target LLM to generate training data for our auxiliary model (\cref{sec:prompting-target-llm}).
Secondly, we set calibration targets in a way that does not require access to the target LLM beyond its generated outputs (\cref{sec:setting-calibration-targets}).
Lastly, we train the auxiliary calibrator to predict the target LLM's confidence for a given question (\cref{sec:training-auxiliary-model}).
Thereby, we contribute two parts that are agnostic to the LLM in question: The creation of calibration targets and their prediction through the auxiliary model.
Note that we will use the terms auxiliary model or calibrator interchangeably in the following sections.

\subsection{Prompting the Target LLM}\label{sec:prompting-target-llm}

\begin{figure}[tb!]
    \centering
    \begin{subfigure}[b]{\columnwidth}
        \centering
        \includegraphics[width=0.85\textwidth]{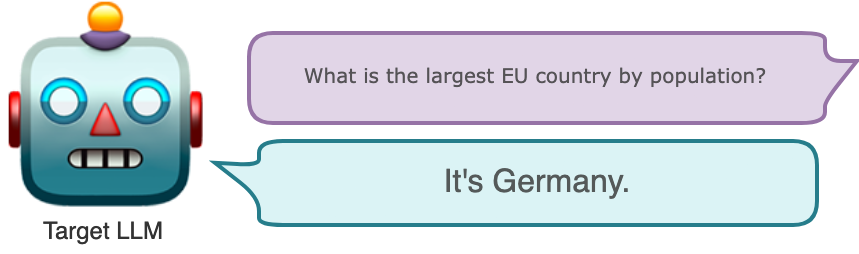}
        \caption{Default prompting.}
        \label{subfig:default-prompting}
    \end{subfigure}
    \par\bigskip
    \begin{subfigure}[b]{\columnwidth}
        \centering
        \includegraphics[width=0.85\textwidth]{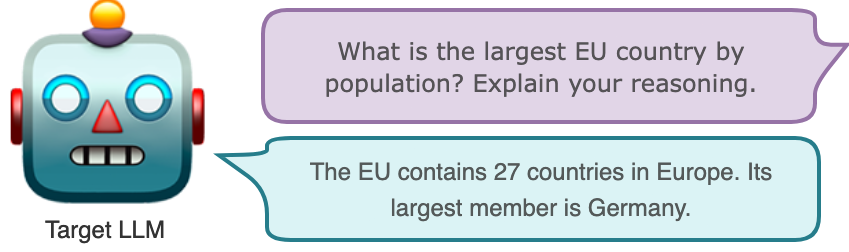}
        \caption{Chain-of-though prompting.}
        \label{subfig:cot-prompting}
    \end{subfigure}
    \par\bigskip
    \begin{subfigure}[b]{\columnwidth}
        \centering
        \includegraphics[width=0.85\textwidth]{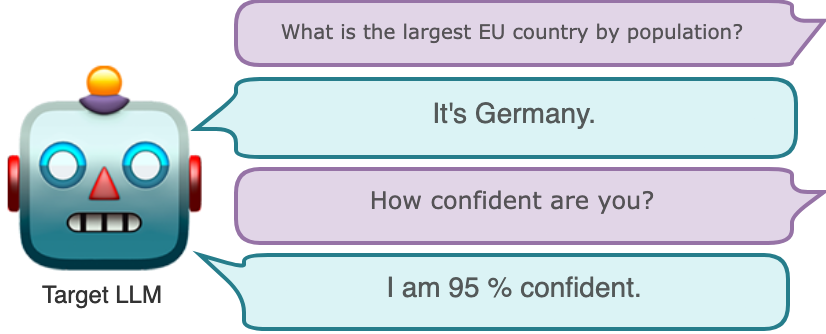}
        \caption{Prompting with verbalized confidence.}
        \label{subfig:verbalized-promptng}
    \end{subfigure}
    \caption{Illustration of the prompting strategies used to generate the input data for the auxiliary calibrator. Note that (c) can also involve confidence expressed in words (``My confidence level is low'') and that (b) and (c) can be combined. The exact prompts are listed in \cref{app:prompting}.}\label{fig:prompting-methods}
\end{figure}

In this step, we generate finetuning data for the auxiliary model by prompting the target LLM on the given task. 
Here, we explore different variations to see which model response might provide the best training signal for the auxiliary calibrator.
More concretely, while the original prompt and model generation might already suffice to predict the model's confidence, we also ask the model to elaborate on its answer using chain-of-thought prompting \citep{wei2022chain}.
We hypothesize that including additional reasoning steps could expose signals that are useful for the calibrator.\footnote{We do this while acknowledging evidence by \citet{turpin2023language} that shows that any chain-of-thought reasoning might not reflect the actual reasons for a specific model response.}
We furthermore take a model's assessment of its confidence into account, too.
Recent works on \emph{verbalized uncertainty} \citep{lin2022teaching, tian2023just}  investigated how to elicit such an assessment as a percentage value, e.g.\@ ``I am 95 \% confident in my answer'', or using linguistic expressions such as ``My confidence is somewhat low''.
While previous studies like \citet{zhou2024relying} have demonstrated the difficulty in obtaining reliable self-assessments, we can just treat them as additional input features, and let their importance be determined through the auxiliary model training.
We illustrate the different prompting strategies in \cref{fig:prompting-methods} and list the exact prompts in \cref{app:prompting}.

\subsection{Setting Calibration Targets}\label{sec:setting-calibration-targets}

After explaining the inputs to the auxiliary model, the question naturally arises about what the calibrator should be trained to predict.
The work by \citet{mielke2022reducing} introduces an additional model that simply predicts individual answer correctness (and does so by using the target model's internal hidden states, which is not possible for black-box models).
While we also check test this type of output in \cref{sec:calibration-experiments}), we show that we can produce better calibration targets through clustering.

\paragraph{Background.} We consider the notion of calibration by \citet{guo2017calibration}:
We define a predictor as \emph{calibrated} when given some predicted outcome $\hat{y} \in \mathcal{Y}$ and some associated probability $\hat{p} \in [0, 1]$, we have
\begin{equation}\label{eq:calibration}
    \mathbb{P}\big(Y = \hat{y}\ |\ P = \hat{p} \big) = \hat{p},
\end{equation}
\noindent i.e.\@ the probability $\hat{p}$ corresponding to the actual relative frequency of correct predictions.
This quantity is often approximated empirically through means such as the expected calibration error (ECE; \citealp{naeini2015obtaining}).
It evaluates the error
\begin{equation}\label{eq:calibration-error}
    \mathbb{E}\Big[\big|\mathbb{P}\big(Y = \hat{y}\ |\ P = \hat{p}  \big) - \hat{p}\big|\Big],
\end{equation}
where in the case of the ECE, the expectation is computed by grouping $N$ test predictions into $M$ equally wide bins. 
Defining $\mathcal{B}_m$ as the set indices that belong to bin $m$, we can write the ECE as 
\begin{equation}\label{eq:ece}
    \sum_{m=1}^M \frac{|\mathcal{B}_m|}{N}\Big|\underbrace{\frac{1}{|\mathcal{B}_m|}\sum_{i \in \mathcal{B}_m}\mathbf{1}(\hat{y}_i = y_i)}_{\text{Bin accuracy (target)}} - \underbrace{\frac{1}{|\mathcal{B}_m|}\sum_{i \in \mathcal{B}_m}\hat{p}_i}_{\text{Avg. bin confidence}}\Big|, 
\end{equation}
\noindent where $\mathbf{1}(\hat{y}_i = y_i)$ is the indicator function showing whether the prediction was correct.
As \citet{guo2017calibration} note, both terms in the difference approximate the left-hand and right-hand side in \cref{eq:calibration} per bin, respectively.


\paragraph{Contribution.} Our key insight is here that we can optimize an objective similar to \cref{eq:calibration-error}, inspired by the way that the ECE in \cref{eq:ece}  aggregates samples in homogeneous groups and measures the group-wise accuracy.
This done without changing the LLM's original answers or access to token probabilities.
Instead of creating bins $\mathcal{B}_m$ by confidence, which is not possible in a black-box setting, we create clustered sets $\mathcal{C}_m$ of inputs with similar sentence embeddings.
Calibration targets are then obtained by using the observed accuracy per set $\mathcal{C}_m$.
This is similar to \citet{lin2022teaching}, who consider the accuracy per question category.
Yet in the absence of such metadata, we expect good embedding and clustering algorithms to roughly group inputs by category.
\citet{holtgen2023richness} also echo a similar sentiment, describing how ECE's grouping by confidence can be abstracted to other kinds of similarities.
They also provide a proof that the calibration error of a predictor based on a $k$-nearest neighbor clustering tends to zero in the infinite data limit.

Practically, we embed questions into a latent space using a light-weight model such as SentenceBERT \citep{reimer2019sentence}, normalize the embeddings along the feature dimension \citep{timkey2021all}, and then use HDBSCAN \citep{campello2013density} to cluster them into questions of similar topic. 
The use of HDBSCAN has multiple advantages: 
Compared to e.g.\@ $k$-means, we do not have to determine the numbers of clusters in advance, and since the clustering is conducted bottom-up, clusters are not constrained to a spherical shape. 
Furthermore, compared to its predecessor DBSCAN \citep{ester1996density}, HDBSCAN does not require one to determine the minimum distance between points for clustering manually.
We evaluate this procedure in \cref{sec:clustering-eval,app:additional-clustering}.



\subsection{Training the Auxiliary Model}\label{sec:training-auxiliary-model}

After determining the input and the training targets for the auxiliary model in the previous sections, we can now describe the actual training procedure that makes it predict the target LLM's confidence.
To start, we feed the questions alongside some in-context samples into our target LLM.
We retain the generated answers and create a dataset that combines the question (without in-context samples) and the target model's answers.
These are used to train the auxiliary calibrator to predict the calibration targets obtained by the clustering procedure above.
In our experiments, we use DeBERTaV3 \citep{he2023debertav3}, an improvement on the original DeBERTa model \citep{he2021deberta} using ELECTRA-style pre-training \citep{clark2020electra} and other improvements.
We then finetune it using the AdamW optimizer \citep{loshchilov2018fixing} in combination with a cosine learning rate schedule.
We minimize the following mean squared error, where $\hat{p}_i$ is the predicted confidence, $\mathcal{C}(i)$ the cluster that the input question with index $i$ belongs to and $\hat{a}_j$ an answer given by the target LLM: 
\begin{equation}
\Big(\hat{p}_i - \underbrace{\frac{1}{|\mathcal{C}(i)|}\sum_{j \in \mathcal{C}(i)}\mathbf{1}(\hat{a}_j\text{ is correct})}_{\text{Cluster accuracy (target)}}\Big)^2.
\end{equation}
We also explore a variant that simply predicts whether the LLM's answer is expected to be correct or incorrect.
In this case, we optimize a binary cross-entropy loss with loss weights.\footnote{The loss weights are based on \texttt{scikit-learn}'s implementation using the ``balanced'' mode, see \url{https://scikit-learn.org/stable/modules/generated/sklearn.utils.class_weight.compute_class_weight.html}.}
Finally, we select the final model via the best loss on the validation set.
We determine the learning rate and weight decay term through Bayesian hyperparameter search \citep{snoek2012practical}, picking the best configuration by validation loss.
We detail search ranges and found values in \cref{app:hyperparameters}.
Training hardware and the environmental impact are discussed in \cref{app:enviromental-impact}.

\section{Experiments}\label{sec:experiments}

We now demonstrate how APRICOT \peach\ provides a simple yet effective solution to calibrate LLMs.
Before assessing the quality of the unsupervised clustering to determine calibration targets from \cref{sec:methodology}, we first introduce the dataset and models.

\paragraph{Datasets.} We employ TriviaQA \citep{joshi2017triviaqa}, a common (closed-book) question-answering dataset.
Open-ended question answering is an ideal testbed for natural language generation tasks, since it is comparatively easy to check whether an answer is correct or not, so calibration has an intuitive interpretation.
To preprocess TriviaQA, we create a training set of 12k examples and choose another 1.5k samples as a validation and test split, respectively.\footnote{Since the original test split does not include answers, we generate the validation and test split from the original validation split.} 
Secondly, we run experiments on CoQA \citep{reddy2019coqa}, a conversational question-answering dataset in which the model is quizzed about the information in a passage of text.
We treat the dataset as an open-book dataset, where the model is shown the passage and then asked one of the corresponding questions at a time.
We extract a subset of the dataset to match the split sizes of TriviaQA.

\paragraph{Models.} For our white-box model experiments, we choose a 7 billion parameter variant of the Vicuna v1.5 model \citep{zheng2023judging},\footnote{\url{https://huggingface.co/lmsys/vicuna-7b-v1.5}.}
an instruction-finetuned model originating from Llama 2 \citep{touvron2023llama}.
For the black-box model, we opt for OpenAI's GPT-3.5 \citep{chatgpt}.\footnote{Specifically, using version \texttt{gpt-3.5-turbo-0125}.}
Despite recent API changes granting access to token probabilities,\footnote{\url{https://x.com/OpenAIDevs/status/1735730662362189872} (last accessed on 16.01.24).} creating methods for black-box confidence estimation is still relevant for multiple reasons: Token probabilities are not available for most black-box models, they might be removed again to defend against potential security issues; and they are not always a reliable proxy for confidence.

\subsection{Setting Calibration Targets by Clustering}\label{sec:clustering-eval}
Before beginning our main experiments, we would like to verify that our proposed methodology in \cref{sec:setting-calibration-targets} is sound. 
In particular, clustering the embeddings of questions and computing the calibration confidence targets rests on the assumption that similar questions are collected in the same cluster.
Ideally, we would like to check this using metadata, which however is usually not available.

\paragraph{Setup.} Instead, we evaluate this through different means: 
We first use the \texttt{all-mpnet-base-v2} model from the sentence transformers package \citep{reimer2019sentence} and HDBSCAN with a minimum cluster size of $3$ to cluster questions.
We then analyze the textual and semantic similarity of questions in a cluster by computing the average pair-wise ROUGE-L score (\emph{semantic}; \citealp{lin2004rouge})\footnote{As implemented by the \texttt{evaluate} package, see \url{https://huggingface.co/docs/evaluate/index}.} between questions and cosine similarities between question \emph{embeddings} of the same cluster (\emph{semantic}).
Since performing this evaluation on the entire dataset is computationally expensive, we approximate the score by using $5$ pairwise comparisons per cluster, with $200$ comparisons for ROUGE-L and $1000$ for cosine similarity in total, respectively. 
As a control for our method (\emph{clustering}), we also compute values between unrelated questions that are not in the same cluster (\emph{random}).

\begin{table}
    \renewcommand{\arraystretch}{1.5}
    \centering
    \resizebox{\columnwidth}{!}{
    \begin{tabular}{@{}lrrrr@{}}
        \toprule
        & \multicolumn{2}{c}{TriviaQA} & \multicolumn{2}{c}{CoQA}  \\
        \cmidrule(lr){2-3} \cmidrule(lr){4-5}
         & Textual & Semantic & Textual & Semantic \\
        \toprule
        Random & $.11{\scriptstyle\ \pm.08 }$ & $.00{\scriptstyle\ \pm.08}$ & $.08{\scriptstyle\ \pm.12}$ & $.00{\scriptstyle\ \pm.12}$ \\
         Clustering & $.39{\scriptstyle\ \pm.28}$ & $.60{\scriptstyle\ \pm.14}$ & $.47{\scriptstyle\ \pm.25}$ & $.70{\scriptstyle\ \pm.17}$  \\ 
        \bottomrule
    \end{tabular}%
    }
    \caption{Results of evaluation of found clusters on TriviaQA and CoQA, including one standard deviation. 
    }\label{fig:clustering-results}
\end{table}

\paragraph{Results.} We show the results of this analysis in \cref{fig:clustering-results}.
We observe noticeable differences between the random baseline and the similarity for the clustering scores, both on a textual and semantic level.
While there is smaller difference on a textual level due to the relatively similar wording of questions, the semantic similarity based on the encoded questions is very notable. 
We provide deeper analyses of this part in \cref{app:additional-clustering}, showing that this method creates diverse ranges of calibration confidence targets.
This suggests two things: On the one hand, our proposed methodology is able to identify fine-grained categories of questions. On the other hand, the diversity in calibration targets indicates that we detect sets of questions on which the LLM's accuracy varies---and that this variety should be reflected. 
We test the ability of different methods to do exactly this next.


\subsection{Calibrating White- and Black-Box Models}\label{sec:calibration-experiments}

We are interested to see whether auxiliary models can reliably predict the target LLM's confidence.
We describe our experimental conditions below.

\paragraph{Evaluation metrics.} Aside from reporting the accuracy on the question answering task, we also report several calibration metrics, including the expected calibration error (ECE; \citealp{naeini2015obtaining}) using $10$ bins.
In order to address any distortion of results introduced by the binning procedure, we use smooth ECE (smECE; \citealp{blasiok2023smooth}), which avoids the binning altogether by smoothing observations using a RBF kernel. 
We also consider Brier score \citep{brier1950verification}, which can be interpreted as mean-squared error for probabilistic predictions.
We further show how indicative the predicted confidence is for answering a question incorrectly by measuring the AUROC.
The AUROC treats the problem as a binary misprediction detection task based on the confidence scores, aggregating the results over all possible decision thresholds.
In each case, we report the result alongside a bootstrap estimate of the standard error \citep{efron1994introduction} estimated from $100$ samples and test for significance using the ASO test \citep{del2018optimal, dror2019deep, ulmer2022deep} with $\tau = 0.35$ and a confidence level of $\alpha = 0.1$.

\paragraph{Baselines.} To contextualize the auxiliary calibrator results, we consider the following baselines:
We consider the raw (length-normalized) sequence likelihoods (Seq. likelihood) as well as variant using Platt scaling \citep{platt1999probabilistic}:
Using the raw likelihood $\hat{p} \in [0, 1]$ and the sigmoid function $\sigma$, we fit two additional scalars $a, b \in \mathbb{R}$ to minimize the mean squared error on the validation set to produce a calibrated likelihood $\hat{q} = \sigma(a\hat{p} + b)$ while keeping all other calibrator parameters fixed.
We also compare it to the recent method of \emph{verbalized} uncertainty \citep{lin2022teaching, tian2023just, anonymous2024can}, where we ask the model to assess its confidence directly. 
We do this by asking for confidence in percent (Verbalized $\%$) and using a seven-point scale from ``very low'' to ``very high'' and which is mapped back to numeric confidence scores (Verbalized Qual.). 
Where applicable, we also distinguish between baselines with and without chain-of-thought prompting (CoT; \citealp{wei2022chain}).
We detail this scale and corresponding prompts in \cref{app:prompting}.
For our approach, we distinguish between confidence targets obtained through the procedure in \cref{sec:setting-calibration-targets} (clustering) and simply predicting whether the given answer is correct or incorrect (binary).

\begin{table*}[htb]
    \renewcommand{\arraystretch}{1.35}
    \centering
    \resizebox{0.995\textwidth}{!}{
    \begin{tabular}{@{}llcccccccccccc@{}}
        \toprule
         & & \multicolumn{5}{c}{TriviaQA} & \multicolumn{5}{c}{CoQA} \\
        \cmidrule(lr){3-7} \cmidrule(lr){8-12} 
        & Method & Success & Brier$\downarrow$ & ECE$\downarrow$  & smECE$\downarrow$ & AUROC$\uparrow$ & Success & Brier$\downarrow$ & ECE$\downarrow$  & smECE$\downarrow$ & AUROC$\uparrow$  \\ 
        \toprule
        \multirow{10}{*}{\rotatebox{90}{Vicuna v1.5 (white-box)}} & Seq. likelihood & - & $.22{\scriptstyle\ \pm.01}$& $.05{\scriptstyle\ \pm.00}$& $.03{\scriptstyle\ \pm.00}$& $.79{\scriptstyle\ \pm.01}$ & - & $.32{\scriptstyle\ \pm.01}$& $.08{\scriptstyle\ \pm.00}$& $.08{\scriptstyle\ \pm.00}$& $.69{\scriptstyle\ \pm.01}$ \\ 
        & Seq. likelihood (CoT) & - & $.25{\scriptstyle\ \pm.01}$& $.04{\scriptstyle\ \pm.00}$& $.04{\scriptstyle\ \pm.00}$& $.70{\scriptstyle\ \pm.01}$  & - & $.35{\scriptstyle\ \pm.01}$& $.04{\scriptstyle\ \pm.00}$& $.05{\scriptstyle\ \pm.00}$& $.61{\scriptstyle\ \pm.01}$ \\ 
        & Platt scaling & - & $.24{\scriptstyle\ \pm.00}$& $.08{\scriptstyle\ \pm.00}$& $.07{\scriptstyle\ \pm.00}$& $.70{\scriptstyle\ \pm.01}$  & - & $.30{\scriptstyle\ \pm.00}$& $.03{\scriptstyle\ \pm.00}$& $.03{\scriptstyle\ \pm.00}$& $.69{\scriptstyle\ \pm.01}$ \\ 
        & Platt scaling (CoT) & - & $.24{\scriptstyle\ \pm.00}$& $.12{\scriptstyle\ \pm.00}$& $.11{\scriptstyle\ \pm.00}$& $.79{\scriptstyle\ \pm.01}$  & - & $.30{\scriptstyle\ \pm.00}$& $.02{\scriptstyle\ \pm.00}$& $.02{\scriptstyle\ \pm.00}$& $.61{\scriptstyle\ \pm.01}$ \\ 
        & Verbalized Qual. & 0.19 & $.38{\scriptstyle\ \pm.03}$& $.02{\scriptstyle\ \pm.00}$& $.02{\scriptstyle\ \pm.00}$& $.62{\scriptstyle\ \pm.03}$ & $0.66$ & $.45{\scriptstyle\ \pm.01}$& $\underline{\mathbf{.00}}{\scriptstyle\ \pm.00}$& $\underline{\mathbf{.00}}{\scriptstyle\ \pm.00}$& $.48{\scriptstyle\ \pm.01}$ \\ 
         & Verbalized Qual. (CoT) & 0.25 & $.39{\scriptstyle\ \pm.02}$& $\underline{\mathbf{.01}}{\scriptstyle\ \pm.00}$& $\underline{\mathbf{.01}}{\scriptstyle\ \pm.00}$& $.60{\scriptstyle\ \pm.02}$ & $0.73$ & $.45{\scriptstyle\ \pm.01}$& $\underline{\mathbf{.00}}{\scriptstyle\ \pm.00}$& $\underline{\mathbf{.00}}{\scriptstyle\ \pm.00}$& $.48{\scriptstyle\ \pm.01}$\\ 
        & Verbalized $\%$ & 1.00 & $.39{\scriptstyle\ \pm.01}$& $.38{\scriptstyle\ \pm.00}$& $.27{\scriptstyle\ \pm.00}$& $.52{\scriptstyle\ \pm.01}$ & $0.99$ & $.49{\scriptstyle\ \pm.01}$& $.48{\scriptstyle\ \pm.00}$& $.32{\scriptstyle\ \pm.00}$& $.53{\scriptstyle\ \pm.01}$ \\ 
        & Verbalized $\%$ (CoT) & 1.00 & $.39{\scriptstyle\ \pm.01}$& $.38{\scriptstyle\ \pm.00}$& $.26{\scriptstyle\ \pm.00}$& $.49{\scriptstyle\ \pm.01}$ & $0.99$ & $.48{\scriptstyle\ \pm.01}$& $.06{\scriptstyle\ \pm.00}$& $.06{\scriptstyle\ \pm.00}$& $.55{\scriptstyle\ \pm.01}$ \\ 
        \cdashline{2-12}
        & Auxiliary (binary)  & - & $.20{\scriptstyle\ \pm.01}$& $.16{\scriptstyle\ \pm.01}$& $.15{\scriptstyle\ \pm.01}$& $.81{\scriptstyle\ \pm.01}$ &  - & $.20{\scriptstyle\ \pm.01}$& $.16{\scriptstyle\ \pm.01}$& $.15{\scriptstyle\ \pm.01}$& $\underline{\mathbf{.82}}{\scriptstyle\ \pm.01}$ \\ 
        & Auxiliary (clustering) & - & $\underline{\mathbf{.18}}{\scriptstyle\ \pm.00}$& $.09{\scriptstyle\ \pm.01}$& $.09{\scriptstyle\ \pm.01}$& $\underline{\mathbf{.83}}{\scriptstyle\ \pm.01}$ & - & $\underline{\mathbf{.18}}{\scriptstyle\ \pm.00}$& $.04{\scriptstyle\ \pm.01}$& $.04{\scriptstyle\ \pm.01}$& $\mathbf{.82}{\scriptstyle\ \pm.01}$ \\ 
        \midrule
        \multirow{10}{*}{\rotatebox{90}{GPT-3.5 (black-box)}} & Seq. likelihood & - & $.15{\scriptstyle\ \pm.01}$& $.04{\scriptstyle\ \pm.00}$& $.04{\scriptstyle\ \pm.00}$& $.69{\scriptstyle\ \pm.02}$ & - & $.29{\scriptstyle\ \pm.01}$& $.11{\scriptstyle\ \pm.00}$& $.11{\scriptstyle\ \pm.00}$& $.70{\scriptstyle\ \pm.01}$ \\
        & Seq. likelihood (CoT) & - & $.14{\scriptstyle\ \pm.00}$& $.05{\scriptstyle\ \pm.00}$& $.05{\scriptstyle\ \pm.00}$& $.60{\scriptstyle\ \pm.02}$ & - & $.25{\scriptstyle\ \pm.00}$& $\underline{\mathbf{.01}}{\scriptstyle\ \pm.00}$& $\underline{\mathbf{.02}}{\scriptstyle\ \pm.00}$& $.52{\scriptstyle\ \pm.02}$\\
        & Platt scaling & - & $.15{\scriptstyle\ \pm.00}$& $.04{\scriptstyle\ \pm.00}$& $.04{\scriptstyle\ \pm.00}$& $.69{\scriptstyle\ \pm.02}$ & - & $.26{\scriptstyle\ \pm.01}$& $.03{\scriptstyle\ \pm.00}$& $.03{\scriptstyle\ \pm.00}$& $.70{\scriptstyle\ \pm.01}$ \\
        & Platt scaling (CoT) & - & $.15{\scriptstyle\ \pm.00}$& $.12{\scriptstyle\ \pm.00}$& $.12{\scriptstyle\ \pm.00}$& $.60{\scriptstyle\ \pm.02}$ & - & $.25{\scriptstyle\ \pm.00}$& $.06{\scriptstyle\ \pm.00}$& $.06{\scriptstyle\ \pm.00}$& $.52{\scriptstyle\ \pm.02}$  \\
        & Verbalized Qual. & $1.00$ & $.14{\scriptstyle\ \pm.01}$& $.07{\scriptstyle\ \pm.00}$& $.04{\scriptstyle\ \pm.00}$& $.61{\scriptstyle\ \pm.02}$ & 1.00 & $.27{\scriptstyle\ \pm.00}$& $.07{\scriptstyle\ \pm.00}$& $.05{\scriptstyle\ \pm.00}$& $.52{\scriptstyle\ \pm.01}$ \\ 
        & Verbalized Qual. (CoT) & $1.00$ &  $.15{\scriptstyle\ \pm.00}$& $.04{\scriptstyle\ \pm.00}$& $.03{\scriptstyle\ \pm.00}$& $.63{\scriptstyle\ \pm.02}$ & $1.00$ & $.30{\scriptstyle\ \pm.01}$& $.08{\scriptstyle\ \pm.01}$& $.04{\scriptstyle\ \pm.00}$& $.50{\scriptstyle\ \pm.01}$ \\ 
        & Verbalized $\%$ & 1.00 & $.13{\scriptstyle\ \pm.01}$& $.01{\scriptstyle\ \pm.00}$& $\underline{\mathbf{.01}}{\scriptstyle\ \pm.00}$& $.63{\scriptstyle\ \pm.02}$ & $1.00$ & $.34{\scriptstyle\ \pm.01}$& $.25{\scriptstyle\ \pm.00}$& $.22{\scriptstyle\ \pm.00}$& $.54{\scriptstyle\ \pm.01}$ \\ 
        & Verbalized $\%$ (CoT) & 0.99 & $.13{\scriptstyle\ \pm.01}$& $\underline{\mathbf{.00}}{\scriptstyle\ \pm.00}$& $\mathbf{.01}{\scriptstyle\ \pm.00}$& $.63{\scriptstyle\ \pm.02}$ & $0.58$ & $.37{\scriptstyle\ \pm.01}$& $.09{\scriptstyle\ \pm.01}$& $.06{\scriptstyle\ \pm.00}$& $.49{\scriptstyle\ \pm.02}$ \\ 
        \cdashline{2-12}
        & Auxiliary (binary) & - & $.14{\scriptstyle\ \pm.00}$& $.14{\scriptstyle\ \pm.01}$& $.14{\scriptstyle\ \pm.01}$& $.65{\scriptstyle\ \pm.02}$ & - & $.19{\scriptstyle\ \pm.01}$& $.13{\scriptstyle\ \pm.01}$& $.13{\scriptstyle\ \pm.01}$& $\mathbf{.81}{\scriptstyle\ \pm.01}$ \\
        & Auxiliary (clustering) & - & $\underline{\mathbf{.12}}{\scriptstyle\ \pm.01}$& $.06{\scriptstyle\ \pm.01}$& $.06{\scriptstyle\ \pm.01}$& $\underline{\mathbf{.72}}{\scriptstyle\ \pm.02}$ & - & $\underline{\mathbf{.18}}{\scriptstyle\ \pm.00}$& $.02{\scriptstyle\ \pm.01}$& $\mathbf{.02}{\scriptstyle\ \pm.00}$& $\mathbf{.81}{\scriptstyle\ \pm.01}$ \\
        \bottomrule
    \end{tabular}%
    }
    \caption{Calibration results for Vicuna v1.5 and GPT-3.5 on TriviaQA and CoQA. We bold the best results per dataset and model, and underline those that are statistically significant compared to all other results assessed via the ASO test. Results are reported along with a bootstrap estimate of the standard error.
    }\label{tab:calibration-results}
\end{table*}

\begin{figure*}[htb]
    \centering
    \begin{subfigure}[t]{0.5\columnwidth}
        \centering
        \includegraphics[width=\textwidth]{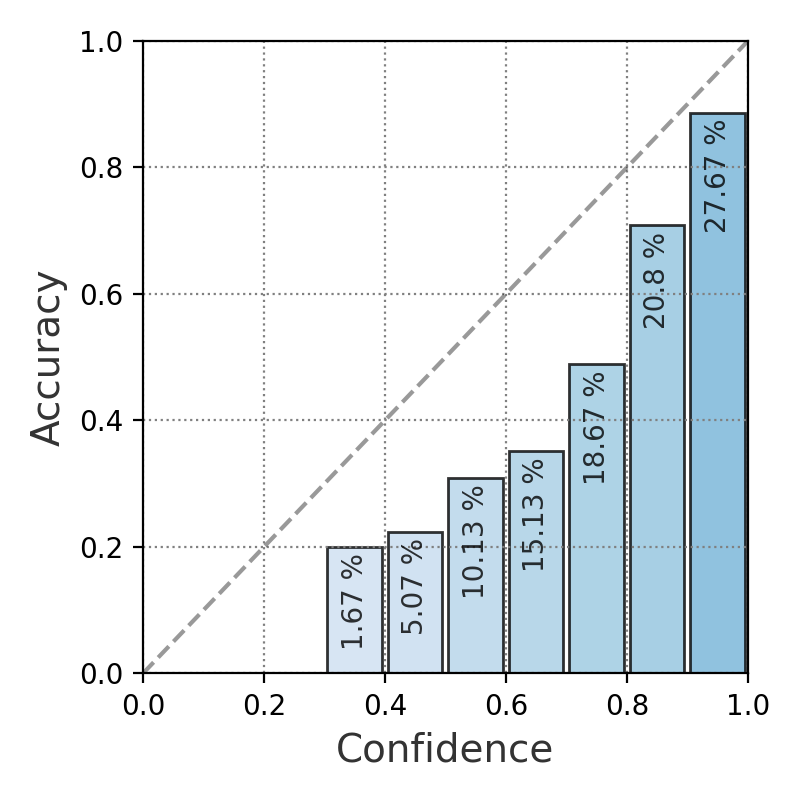}
        \caption{Seq. likelihood.}
        \label{subfig:seq-likelihood}
    \end{subfigure}
    \hfill
    \begin{subfigure}[t]{0.5\columnwidth}
        \centering
        \includegraphics[width=\textwidth]{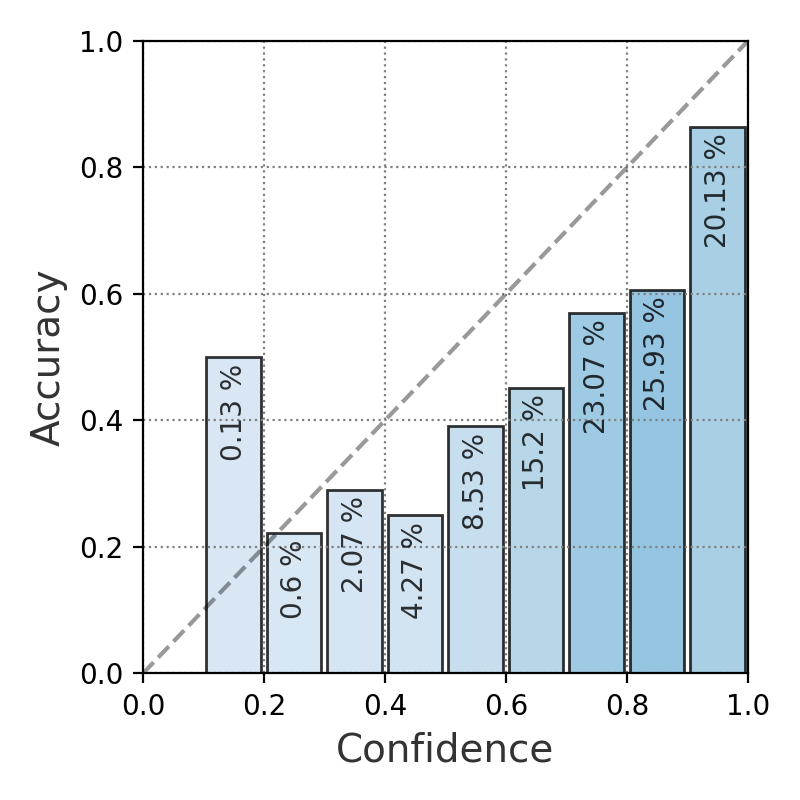}
        \caption{Seq. likelihood (CoT).}
        \label{subfig:temperature-scaling}
    \end{subfigure}
    \hfill
    \begin{subfigure}[t]{0.5\columnwidth}
        \centering
        \includegraphics[width=\textwidth]{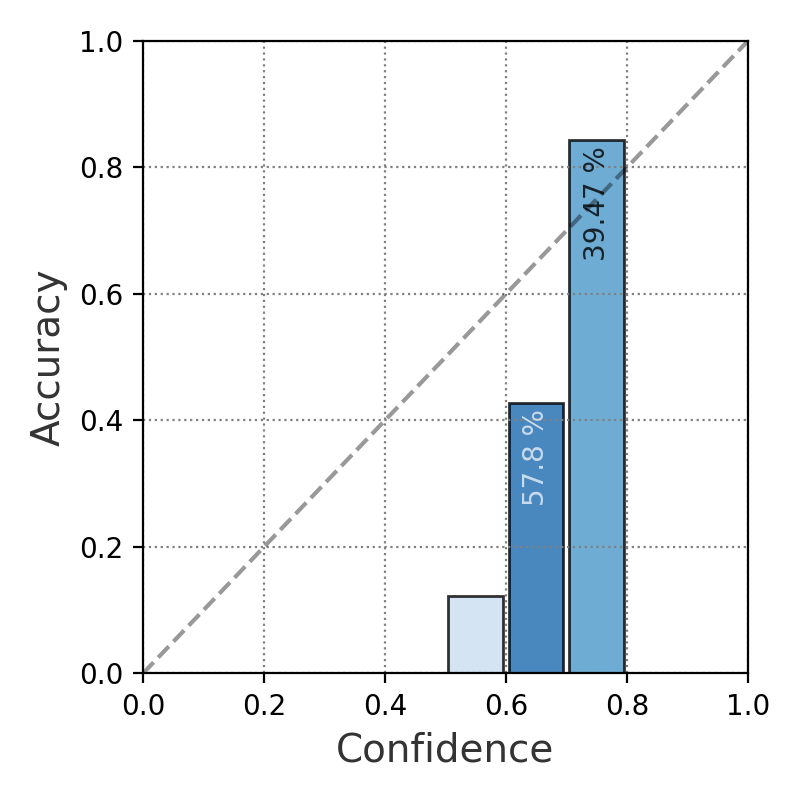}
        \caption{Platt scaling.}
        \label{subfig:verbalized-percentage}
    \end{subfigure}
    \hfill
    \begin{subfigure}[t]{0.5\columnwidth}
        \centering
        \includegraphics[width=\textwidth]{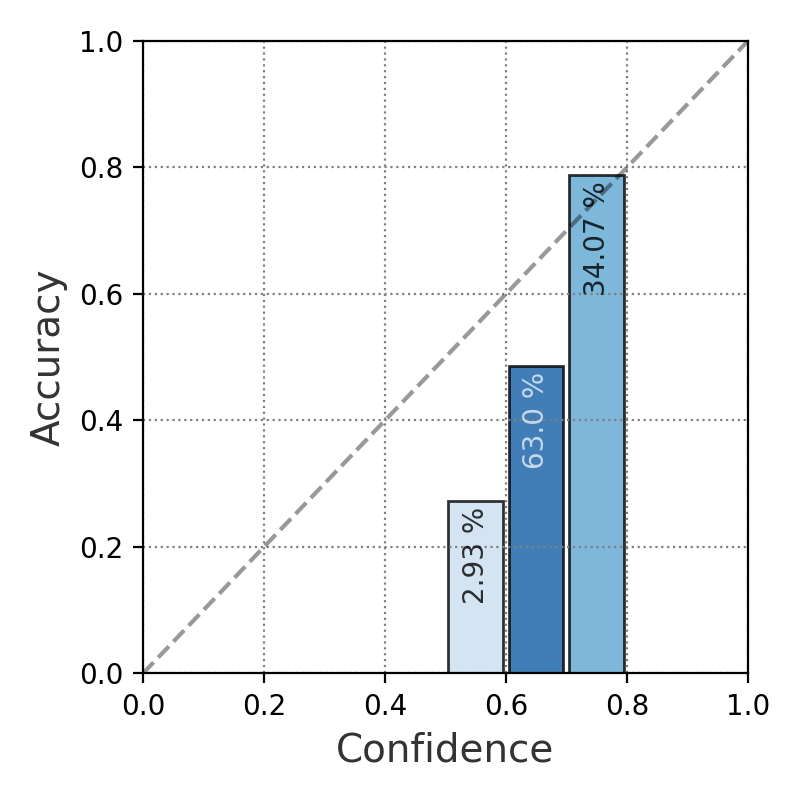}
        \caption{Platt scaling (CoT).}
        \label{subfig:verbalized-qualitative}
    \end{subfigure}\\
    \begin{subfigure}[t]{0.5\columnwidth}
        \centering
        \includegraphics[width=\textwidth]{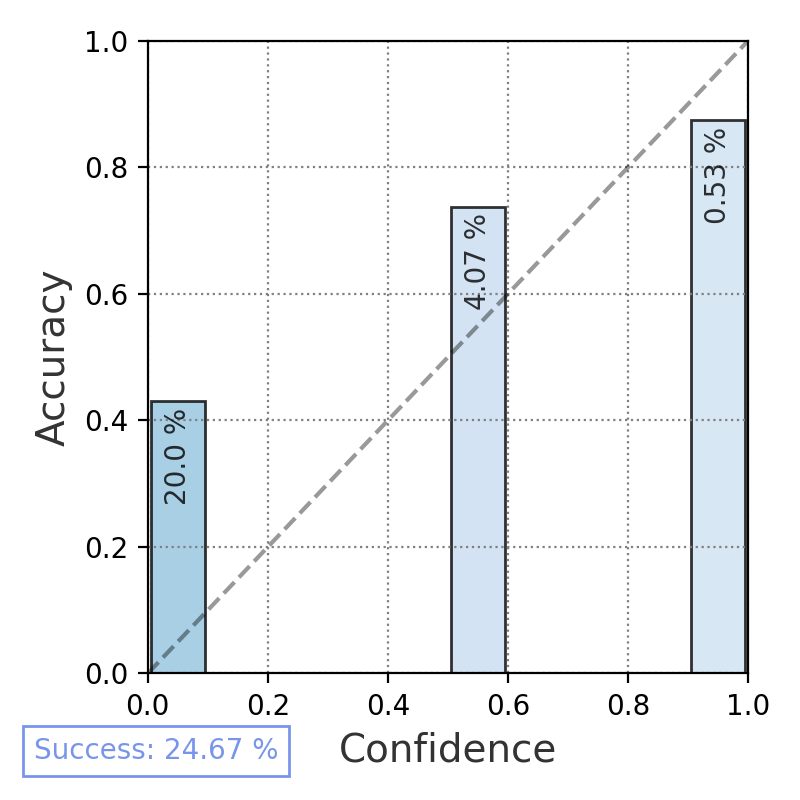}
        \caption{Verbalized Qual.}
        \label{subfig:seq-likelihood}
    \end{subfigure}
    \hfill
    \begin{subfigure}[t]{0.5\columnwidth}
        \centering
        \includegraphics[width=\textwidth]{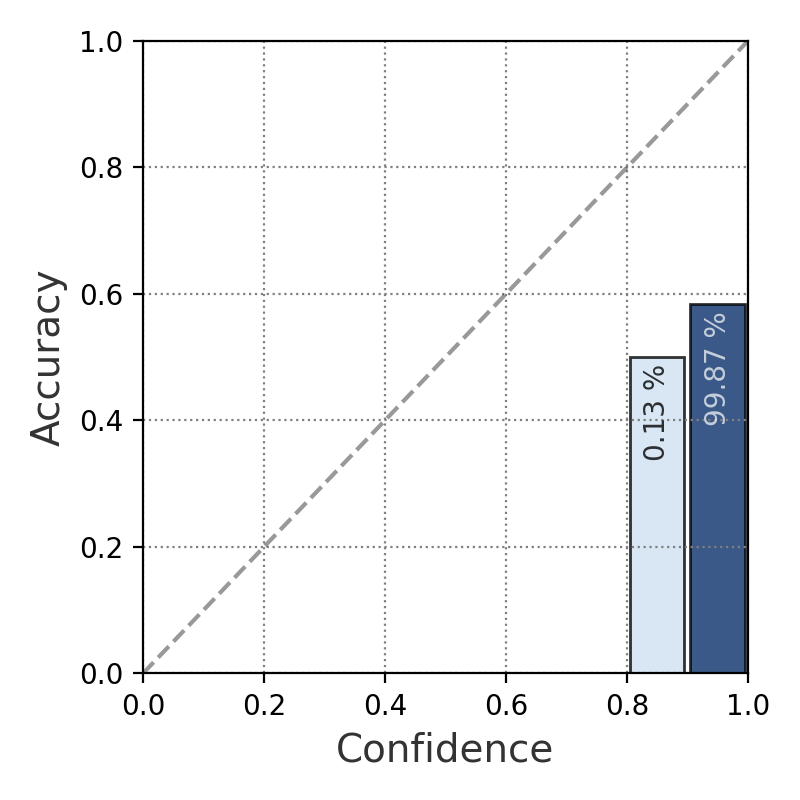}
        \caption{Verbalized $\%$.}
        \label{subfig:temperature-scaling}
    \end{subfigure}
    \hfill
    \begin{subfigure}[t]{0.5\columnwidth}
        \centering
        \includegraphics[width=\textwidth]{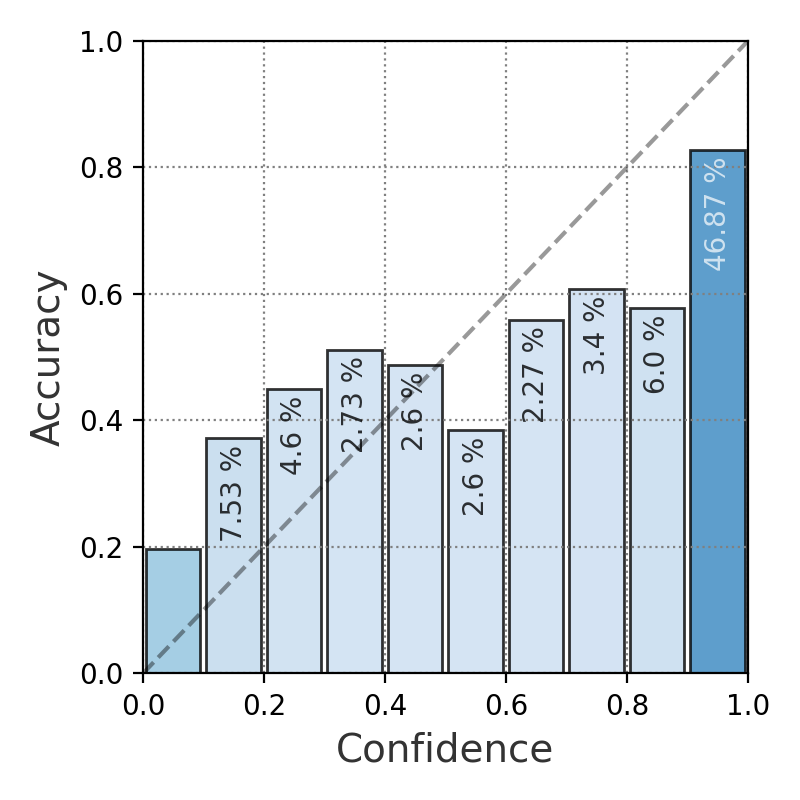}
        \caption{Auxiliary (binary).}
        \label{subfig:verbalized-percentage}
    \end{subfigure}
    \hfill
    \begin{subfigure}[t]{0.5\columnwidth}
        \centering
        \includegraphics[width=\textwidth]{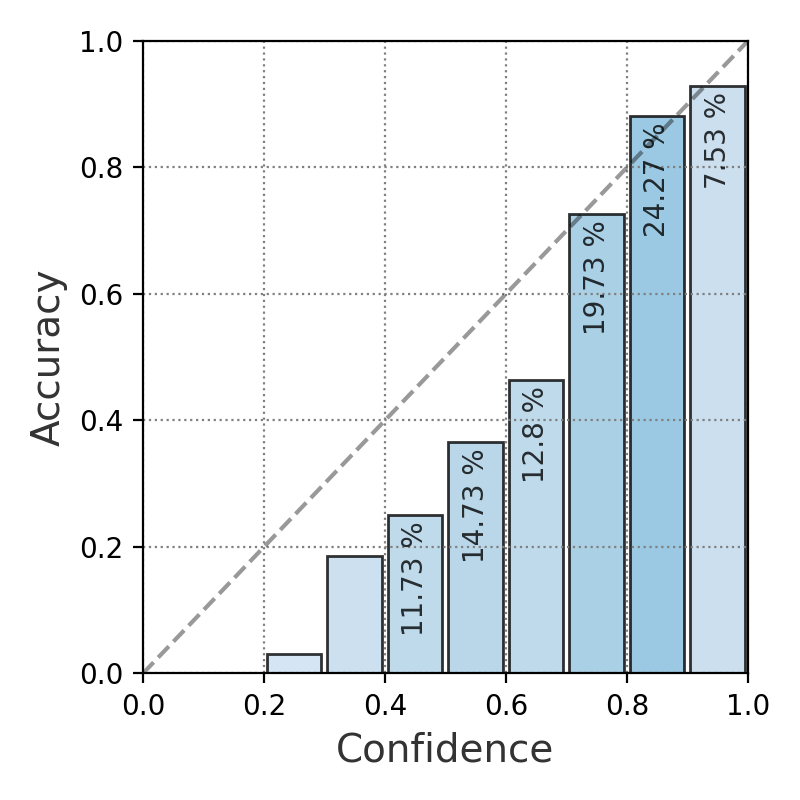}
        \caption{Auxiliary (clustering).}
        \label{subfig:verbalized-qualitative}
    \end{subfigure}
    \caption{Reliability diagrams for our different methods using $10$ bins each for Vicuna v1.5 on TriviaQA. The color as well as the percentage number within each bar indicate the proportion of total points contained in each bin.}\label{fig:reliabiliy-diagrams-vicuna-trivia-qa}
\end{figure*}

\paragraph{Results.} Vicuna v1.5 7B achieves $58 \%$ accuracy on TriviaQA and $44 \%$ on CoQA, while GPT-3.5 obtains $85 \%$ and $55 \%$ accuracy, respectively.\footnote{We describe the evaluation protocol in \cref{app:evaluation-protocol}. Since GPT-3.5 is a closed-source model, it is hard to say whether the higher accuracy scores are due to better model quality, test data leakage, or overlap in questions in the case of TriviaQA \citep{merlo2021question}.} 
We present the calibration results in \cref{tab:calibration-results}. 
APRICOT \peach\ hereby achieves the highest AUROC in all settings and among the lowest Brier scores and calibration errors.
On the latter metric, verbalized confidence beats our method, but often at the cost of a higher worst-case calibration error and lower AUROC. 
In addition, the qualitative verbalized uncertainty does not work reliably for the smaller Vicuna v1.5 model.
CoT prompting seems to increase the success rate of verbalized uncertainty. The additional results on GPT-3.5 suggests that this ability might also be dependent on model size.
The effect of CoT prompting on calibration, however, remains inconsistent across different baselines.
While verbalized uncertainties often perform well according to calibration error, these results have to be taken with a grain of salt: 
Especially for the relatively small 7B Vicuna v1.5 model, the generations do not always contain the desired confidence expression, as visible by the low success rate.
And even when taking the generated confidence expression, their ability to distinguish potentially correct from incorrect LLM responses remains at or close to random level.
Lastly, APRICOT \peach\ with clustering beats the use of binary targets for Vicuna v1.5 and GPT-3.5 on both TriviaQA and CoQA.
We also juxtapose reliability diagrams for the different methods for Vicuna v1.5 on TriviaQA in \cref{fig:reliabiliy-diagrams-vicuna-trivia-qa} (we show the other reliability diagrams, including for GPT-3.5, in \cref{app:additional-calibration}).
Here it becomes clear that verbalized uncertainties approaches usually do not emit a wide variety of confidence scores. 
This is in line with observations by \citet{zhou2023navigating}, who hypothesize the distribution of expressions generated by verbalized uncertainty heavily depend on the mention of e.g. percentage values in the model's training data. 
While \cref{fig:reliabiliy-diagrams-gpt35-trivia-qa-full} shows that GPT-3.5 provides more variety in this regard, the overall phenomenon persists.
We now conduct some additional analyses based on the clustering-based variant of our method.

\subsection{What does the calibrator learn from?}

The previous results pose the question of which parts of input the auxiliary model actually learns from.
So, analogous to the different prompting strategies in \cref{fig:prompting-methods}, we explore different input variants:
First, we test question-only, where the target LLM's answer is omitted completely.
We also test the performance of the calibrator when given more information, for instance the model answer with and without chain-of-thought prompting, which could potentially expose flaws in the LLM's response.\footnote{Based on the recent study by \citet{turpin2023language}, we assume that CoT does \emph{not} expose the LLM's actual reasoning. Nevertheless, it provides more context about the given answer.}
Finally, we also expose the verbalized uncertainty of the LLM to the calibrator.

\paragraph{Results.} We show these results in \cref{tab:calibration-results-features} in \cref{app:additional-calibration}. 
Interestingly, we can observe that even based on the question to the LLM alone, APRICOT \peach\ can already achieve respectable performance across all metrics.
This suggests that the calibrator at least partially learns to infer the difficulty of the LLM answering a question from the type of question alone.
Nevertheless, we also find that adding the LLM's actual answer further improves results, with additional gain when using CoT prompting.
In some cases, the calibration error can be improved when using the LLM's verbalized uncertainties; in this sense, we can interpret the role of the calibrator as mapping the model's own assessment to a calibrated confidence score.

\section{Discussion}

Despite the difficulty of predicting the LLM's confidence from its generated text alone, our experiments have shown that APRICOT \peach\ can be used to produce reasonable scores even under these strict constraints.
We showed in the past sections that the auxiliary model can be finetuned to learn from multiple signals.
On the one hand, the auxiliary calibrator learns a mapping from a latent category of question to the expected difficulty for a target LLM.
On the other hand, including the answer given through CoT prompting and including the LLM's own assessment of its uncertainty helped to further improve results.
While sometimes beaten in terms of calibration error, our method consistently outperforms our baselines in misprediction AUROC, meaning that it can provide the best signal to detect wrong LLM answers.
Compared to other approaches, this yields some desirable properties: 
APRICOT \peach\ is available when sequence likelihood is not; it is more reliable than verbalized uncertainty; and it only needs a light finetuning once, adding negligible inference overhead.
Compared to other methods such as \citet{kuhn2023semantic, lin2023generating}, it also does not require more generations for the same input, reducing the more expensive LLM inference costs.

\section{Conclusion}

We presented APRICOT \peach, a general method to obtain confidence scores from any language model on the input and text output alone. 
We showed that it is possible to compute calibration targets through the clustering of question embeddings.
Through the subsequent finetuning of a smaller language model, we then outperform other methods to distinguish incorrect from correct answers with competitive calibration scores, on different models and datasets.
While we only presented a first, more fundamental version this approach in this work, it lends itself naturally to a whole body of research that aims to improve the calibration of pretrained language models \citep{desai2020calibration, jian2021how, chen2023close}. 
Lastly, future studies might also investigate the uncertainty of the auxiliary model itself and use techniques such as conformal prediction \citep{vovk2005algorithmic, papadopoulos2002inductive, angelopoulos2021gentle} to produce estimates of LLM confidence \emph{intervals}.

\section*{Limitations}

\paragraph{Clustering.} While yielding generally positive results in our case, the clustering methodology from \cref{sec:setting-calibration-targets} requires access to a sufficiently expressive sentence embedding model and a large enough number of data points.
When this is not given, we show that the binary approach---tuning the auxiliary model to predict misprediction---is a viable alternative that can even outperform the clustering variant in some settings.

\paragraph{Distributional shift.} As any neural model, the auxiliary calibrator might be prone to distributional shift and out-of-distribution data.
Further research could help to understand how this issue can be reduced and which parts of the input the model identifies to predict confidence scores in order to unveil potential shortcut learning \citep{du2023shortcut}. 

\paragraph{Other types of language generation.} Our experiments focused on open-ended question answering tasks, which provide an easy way to check answer correctness.
In other types of language generation such as summarization, translation or open text generation, this notion is not given.
However, we point out the relation of our approach to other NLP tasks in \cref{sec:related-work}, which might provide an opening for future research.

\section*{Ethical Considerations}

We mostly see any ethical considerations with our work arising in the general nature of neural models, which therefore also affects our auxiliary calibrator.
The efficacy of neural models might vary on out-of-distribution data.
In safety-sensitive applications, this also means that its predictions might be less trustworthy on certain sub-populations.
In this case, explicit validation of the LLM's answers and the auxiliary calibrators corresponding predictions is necessary and further finetuning might be necessary.
In these cases, reporting the confidence score alongside an answer might also be replaced by withholding the LLM's response, entirely.

\section*{Acknowledgements}

This work was generously supported by the NAVER corporation.

\nocite{langley00}

\bibliography{anthology,custom}
\bibliographystyle{acl_natbib}

\appendix

\section{Appendix}

This appendix is structured as follows:
We list all used prompts in \cref{app:prompting}, briefly explore the evaluation protocal for the QA task in \cref{app:evaluation-protocol} and list hyperparameter search details for replicability in \cref{app:hyperparameters}.
The rest of the appendix is dedicated to additional results, namely for the analysis of the clustering step in \cref{app:additional-clustering} or for the calibration experiments in \cref{app:additional-calibration}.
We also list our compute hardware and its environmental impact in \cref{app:enviromental-impact}.

\subsection{Prompting Setup}\label{app:prompting}

In the following we elaborate on the prompts used in this work.
In general, we use a very simple prompt for question answering, where we fill in a template of the form ``Question: \textcolor{slotcolor}{\{Question\}} Answer:''. 
For in-context samples, we prepend the demonstrations to the input, using the sample template as above.
In the case of chain-of-thought prompting, we use the prompting below:

\begin{tcolorbox}[width=\columnwidth,colback={white},title={\small QA Chain-of-thought prompt},enhanced,attach boxed title to top right={yshift=-3.5mm, xshift=-5mm}, colbacktitle=white, coltitle=black, top=12pt]  
        \small
        Briefly answer the following question by thinking step by step.
        Question: \textcolor{slotcolor}{\{Question\}} Answer:\\
\end{tcolorbox}

In the case of CoQA, we slightly adjust the prompt template to the following:

\begin{tcolorbox}[width=\columnwidth,colback={white},title={\small CoQA Chain-of-thought prompt},enhanced,attach boxed title to top right={yshift=-3.5mm, xshift=-5mm}, colbacktitle=white, coltitle=black, top=12pt]  
        \small 
        Context: \textcolor{slotcolor}{\{Context\}}\\
        Instruction: Briefly answer the following question by thinking step by step.\\
        Question: \textcolor{slotcolor}{\{Question\}}\\
        Answer:\\
\end{tcolorbox}

Note that here the passage that questions are based on is given first, and chain-of-thought prompting is signaled through the ``Instruction'' field. 
When no chain-of-thought prompting is used, this field is omitted.
For the verbalized uncertainty, we use the following prompts, in which case we omit any in-context samples:

\begin{table}[htb]
    \centering
    \renewcommand{\arraystretch}{1.25}
    \resizebox{0.5\columnwidth}{!}{
    \begin{tabular}{lr}
        \toprule
          Expression & Value  \\
        \toprule
          Very low & $0$ \\
          Low & $0.3$ \\
          Somewhat low & $0.45$ \\
          Medium & $0.5$ \\ 
          Somewhat high & $0.65$ \\
          High & $0.7$ \\
          Very high & $1$ \\
          \bottomrule
    \end{tabular}%
    }
    \caption{Mappings between confidence expressions for qualitative uncertainty and numerical confidence scores.}
    \label{tab:confidence-mappings}
\end{table}

\begin{tcolorbox}[width=\columnwidth,colback={white},title={\small Verbalized uncertainty prompt (quantitative)},enhanced,attach boxed title to top right={yshift=-3.5mm, xshift=-5mm}, colbacktitle=white, coltitle=black, top=12pt]  
        \small
        \textcolor{slotcolor}{\{Question\}} \textcolor{slotcolor}{\{Model answer\}}
        Please provide your confidence in the answer only as one of 'Very Low' / 'Low' / 'Somewhat Low' / 'Medium' / 'Somewhat High' / 'High' / 'Very High':\\
\end{tcolorbox}

When mapping these expressions back to probabilities using the mapping in \cref{tab:confidence-mappings}.

\begin{tcolorbox}[width=\columnwidth,colback={white},title={\small Verbalized uncertainty prompt (qualitative)},enhanced,attach boxed title to top right={yshift=-3.5mm, xshift=-5mm}, colbacktitle=white, coltitle=black, top=12pt]  
        \small
        \textcolor{slotcolor}{\{Question\}} \textcolor{slotcolor}{\{Model answer\}} Please provide your confidence in the answer only in percent (0-100 \%):\\
\end{tcolorbox}

We follow \citet{kuhn2023semantic} and use $10$ in-context samples for the original answer, which are randomly sampled from the training set (but in contrast to \citeauthor{kuhn2023semantic}, we sample different examples for each instance). 
When prompting for verbalized uncertainty, we remove these in-context samples.

\subsection{Question-Answering Evaluation Protocol}\label{app:evaluation-protocol}

In order to check model answers automatically, we take inspiration from the evaluation protocol by \citet{kuhn2023semantic}.
They use ROUGE-L \citep{lin2004rouge} to compare the LLM's answer against the reference answer, and consider the answer as correct if the resulting score surpasses $0.3$.
We improve this protocol by adding the following condition: If the gold answer can be found verbatim in the generated answer, the answer is also considered correct.

\subsection{Finetuning Hyperparameters}\label{app:hyperparameters}

We conduct suites of hyperparameter searches per target LLM, dataset and type of calibration targets (binary and clustering) corresponding to the results in \cref{tab:calibration-results}, resulting in eight different suites. 
We then use these found hyperparameters for the results in \cref{tab:calibration-results-features}.

\begin{table*}[htb]
    \centering
    \renewcommand{\arraystretch}{1.5}
    \resizebox{0.675\textwidth}{!}{
    \begin{tabular}{rrllll}
        \toprule
        & & \multicolumn{2}{c}{TriviaQA} & \multicolumn{2}{c}{CoQA} \\
        \cmidrule(lr){3-4} \cmidrule(lr){5-6}
        & & Binary & Clustering & Binary & Clustering \\
        \midrule
        \multirow{2}{*}{Vicuna v1.5} & learning rate & $1.4 \times 10^{-5}$ & $3.37 \times 10^{-5}$ & $9.58 \times 10^{-5}$ & $8.84 \times 10^{-5}$ \\
        & weight decay & $0.03184$ & $0.008936$ & $0.005793$ & $7.42 \times 10^{-4}$  \\
        \midrule
        \multirow{2}{*}{GPT-3.5}  & learning rate & $2.96 \times 10^{-5}$ & $1.62 \times 10^{-5}$ & $5.12 \times 10^{-5}$ & $5.59 \times 10^{-5}$ \\
        & weight decay & $0.01932$ & $0.01362$ & $0.03327$ & $0.03495$ \\
        \bottomrule
    \end{tabular}%
    }
    \caption{Chosen hyperparameters for our model on different datasets and for different calibration targets.}\label{tab:chosen-hyperparameters}
\end{table*}

\paragraph{Search method and ranges.} For the search, we opt for Bayesian hyperparameter search \citep{snoek2012practical} as implemented by Weights \& Biases \citep{wandb}.
We optimize only two hyperparameters: Learning rate and weight decay.
The learning rate is samples from a log-uniform distribution $\log\ \mathcal{U}[1 \times 10^{-5}, 0.01]$ and weight decay from a uniform distribution $\mathcal{U}[1 \times 10^{-4}, 0.05]$ for a total of $50$ runs and $250$ training steps each.
The final hyperparameters selected are given in \cref{tab:chosen-hyperparameters}.

\paragraph{Other hyperparameters.} When obtaining the responses from Vicuna v1.5 7B, we use a batch size of $4$ and generate for a maximum of $50$ tokens and stop generation when the model tries to generate parts of the prompt, such as ``Question:'' / ``Q:'' or ``Answer:'' / ``A:''.
We also use $10$ in-context samples for TriviaQA, but no in-context samples for CoQA. For the auxiliary calibrator, we use a context size of $512$ tokens, batch size of $32$, gradient clipping with a maximum norm of $10$.

\subsection{Additional Clustering Results}\label{app:additional-clustering}

\begin{figure}
    \centering
    \begin{subfigure}[t]{0.98\columnwidth}
        \centering
        \includegraphics[width=0.98\columnwidth]{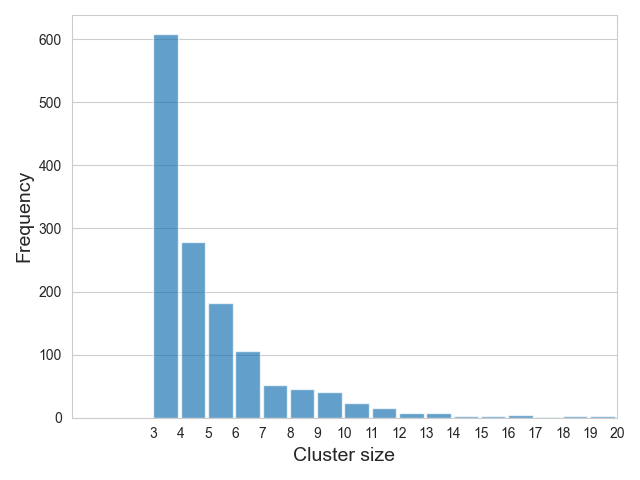}
    \caption{Cluster sizes on TriviaQA.}
    \end{subfigure}
    \vfill
    \begin{subfigure}[t]{0.98\columnwidth}
        \centering
        \includegraphics[width=0.98\columnwidth]{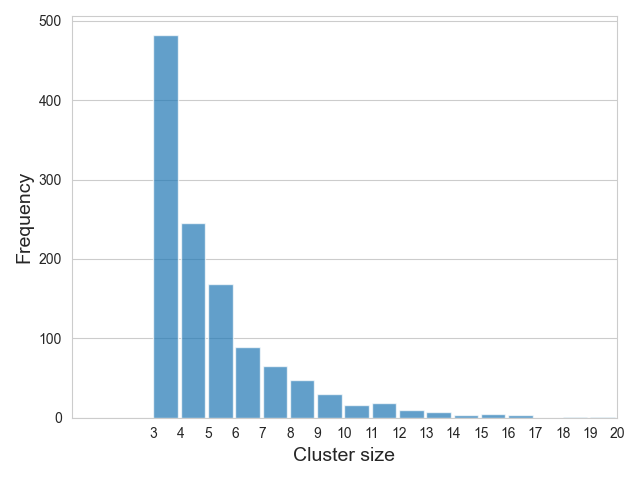}
    \caption{Cluster sizes on CoQA.}
    \end{subfigure}
    \caption{Bar plot of cluster sizes found. The plot is truncated at size 20.}\label{fig:cluster-sizes}
\end{figure}

\begin{table*}
    \centering
     \resizebox{1.975\columnwidth}{!}{
    \begin{tabular}{@{}p{0.95\linewidth}@{}}
    \toprule
    \small Cluster 1120 \\
    \midrule
    \small  How many fluid ounces are in one quarter of an imperial pint? \\
    \small  How many fluid ounces in one Imperial pint? \\
    \small How many fluid ounces in half an Imperial pint? \\
    \toprule
    \small Cluster 920 \\
    \midrule
    \small Which famous US outlaw shot the cashier of a savings bank in Gallatin Missouri in 1869? \\
    \small What famous outlaw committed the Wild West's first train robbery on July 21, 1873 in Adair, Iowa? \\
    \small On July 21, 1873, Jesse James and the James-Younger gang pulled off the first successful what of the American West, in Adair Iowa? \\
    \toprule
    \small Cluster 984 \\
    \midrule
    \small In what country was the game of golf invented?\\
    \small Which ball game was invented by Dr James Naismith in Massachusetts USA  in 1891? \\
    \small It is generally accepted that the game of golf originated in what country? \\
    \small What's a country where most people love playing rugby? \\
    \small What's a country where most people love playing golf? \\
    \toprule
    \small Cluster 64 \\
    \midrule
    \small Which alternative word for the Devil is a Hebrew word with translates as Lord Of The Flies? \\
    \small Beelzebub is Hebrew for what phrase, which is also the title of a famous novel? \\
    \small Diablo is another name for who? \\
    \toprule
    \small Cluster 811 \\
    \midrule
    \small How many colors are there in the spectrum when white light is separated? \\
    \small Which part of the eye contains about 137 million light-sensitive cells in one square inch? \\
    \small Which of the retina's cells can distinguish between different wavelengths of light? \\
    \small In four colour process printing, which is also known as CMYK, which are the only four colours that are used? \\
    \small How many colours are in the rainbow? \\
    \small In art, what are the three primary colours? \\
    \small What color consists of the longest wavelengths of lights visible by the human eye? \\
    \small What are the three primary colours of light? \\
     \toprule
    \small Cluster 74 \\
    \midrule
    \small What was the name of the computer in the movie 2001: A Space Odyssey? \\
    \small What was the name of the computer in Blake's Seven? \\
    \small Developed by IBM, Deep Blue was a computer that played what? \\
    \small What was the name of the PDA produced by Apple, most famous for its handwriting recognition software turning Random House into Condom Nose during a major presentation? \\
    \bottomrule
    \end{tabular}%
    }
    \caption{Contents of some randomly sampled cluster that result from the clustering procedure for TriviaQA.}\label{fig:cluster-contents}
\end{table*}

\begin{table*}
    \centering
     \resizebox{1.975\columnwidth}{!}{
    \begin{tabular}{@{}p{0.95\linewidth}@{}}
    \toprule
    \small Cluster 823 \\
    \midrule
    \small Where in Europe is it located?\\
    \small Is it in the European Plain?\\
    \small Which region of Europe is it in?\\
    \toprule
    \small Cluster 1176 \\
    \midrule
    \small Did she have children?\\
    \small Does she have any children?\\
    \small Did she have any children?\\
    \small Did she have any other children?\\
    \toprule
    \small Cluster 2244 \\
    \midrule
    \small Who won the Kentucky Derby?\\
    \small as he won the Derby before?\\
    \small Has he raced in the Derby before?\\
    \small What were the winning horse's odds?\\
    \small How many Derbys have their been?\\
    \toprule
    \small Cluster 11 \\
    \midrule
    \small Are they densities of everything the same?\\
    \small What is the densest elements at regular conditions?\\
    \small What is density of a substance?\\
    \small What is another symbol for density?\\
    \small Who gives weight per unit volume as the definition?\\
    \small Where is density the same value as it's mass concentration?\\
    \small To make comparisons easier what stands in for density?\\
    \small What is the relative density of something that floats?\\
    \toprule
    \small Cluster 1081 \\
    \midrule
    \small Who was murdered?\\
    \small who was killed?\\
    \small Who committed this murder?\\
    \small who was killed?\\
    \small Who was killed?\\
    \small who was killed?\\
     \toprule
    \small Cluster 579\\
    \midrule
    \small Did it succeed?\\
    \small DId they continue to try?\\
    \small did they succeed?\\
    \small Was it successful?\\
    \small Did they expect that success?\\
    \bottomrule
    \end{tabular}%
    }
    \caption{Contents of some randomly sampled cluster that result from the clustering procedure for CoQA.}\label{fig:cluster-contents-coqa}
\end{table*}

In this section we take a closer look at the results of the clustering procedure described in \cref{sec:setting-calibration-targets}.
In our experiments, we run HDBSCAN using a minimum cluster size of three, since preliminary experiments showed this number to produce the best trade-off between the coherence of cluster contents (as evaluated in \cref{fig:clustering-results}) and a diversity in cluster targets.
This setting yields a distribution of cluster sizes shown in \cref{fig:cluster-sizes}.
We can see that the majority of cluster sizes are rather small, including questions on specific topics, some of which we display in \cref{fig:cluster-contents,fig:cluster-contents-coqa}.
Not shown are cluster sizes over $20$ since the distribution quickly levels off, as well the set of all points that could not be sorted into any cluster.

\begin{figure*}
    \centering
    \begin{subfigure}[t]{0.49\columnwidth}
        \centering
        \includegraphics[width=0.98\columnwidth]{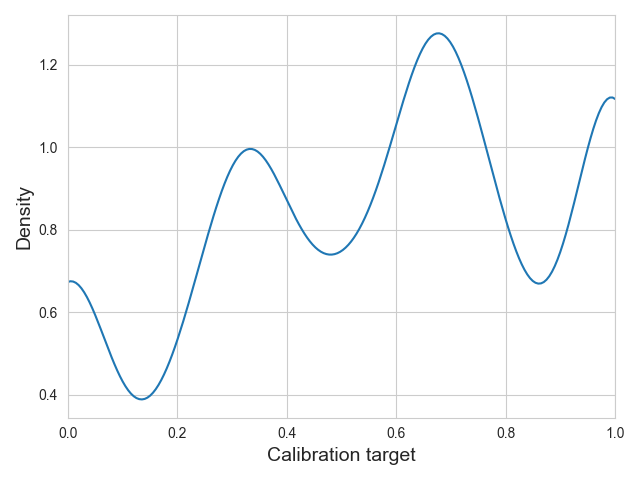}
        \caption{Vicuna v1.5 on TriviaQA.}\label{subfig:calibration-targets-vicuna}
    \end{subfigure}
    \hfill
    \begin{subfigure}[t]{0.49\columnwidth}
        \centering
        \includegraphics[width=0.98\columnwidth]{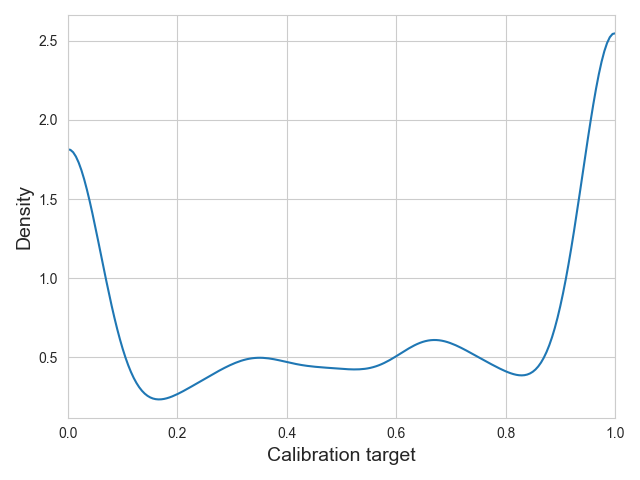}
        \caption{GPT-3.5 on TriviaQA.}\label{subfig:calibration-targets-gpt35}
    \end{subfigure}
    \hfill
    \begin{subfigure}[t]{0.49\columnwidth}
        \centering
        \includegraphics[width=0.98\columnwidth]{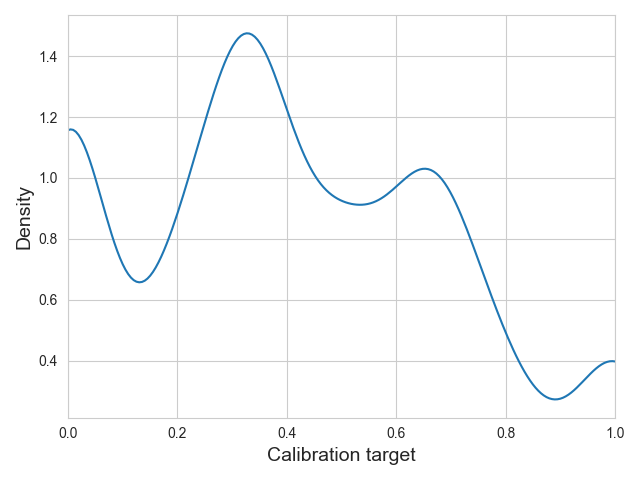}
        \caption{Vicuna v1.5 on CoQA.}\label{subfig:calibration-targets-coqa-vicuna}
    \end{subfigure}
    \hfill
    \begin{subfigure}[t]{0.49\columnwidth}
        \centering
        \includegraphics[width=0.98\columnwidth]{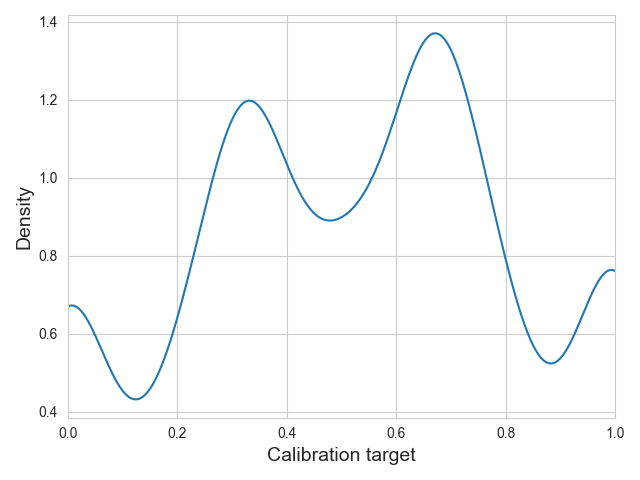}
        \caption{GPT-3.5 on CoQA.}\label{subfig:calibration-targets-coqa-gpt-3.5}
    \end{subfigure}
    \caption{Density plot of calibration targets generated through the clustering procedure for the two LLMs and TriviaQA / CoQA.}\label{fig:calibration-targets}
\end{figure*}

\begin{figure*}
    \centering
    \begin{subfigure}[t]{0.49\columnwidth}
        \centering
        \includegraphics[width=0.98\textwidth]{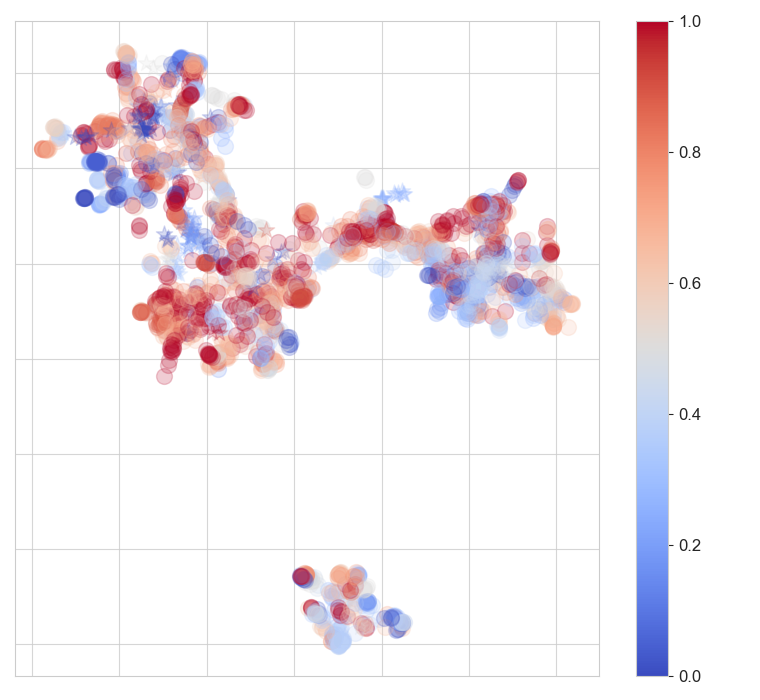}
        \caption{Vicuna v1.5 on TriviaQA.}\label{subfig:clustering-vicuna}
    \end{subfigure}
    \hfill
    \begin{subfigure}[t]{0.49\columnwidth}
        \centering
        \includegraphics[width=0.98\textwidth]{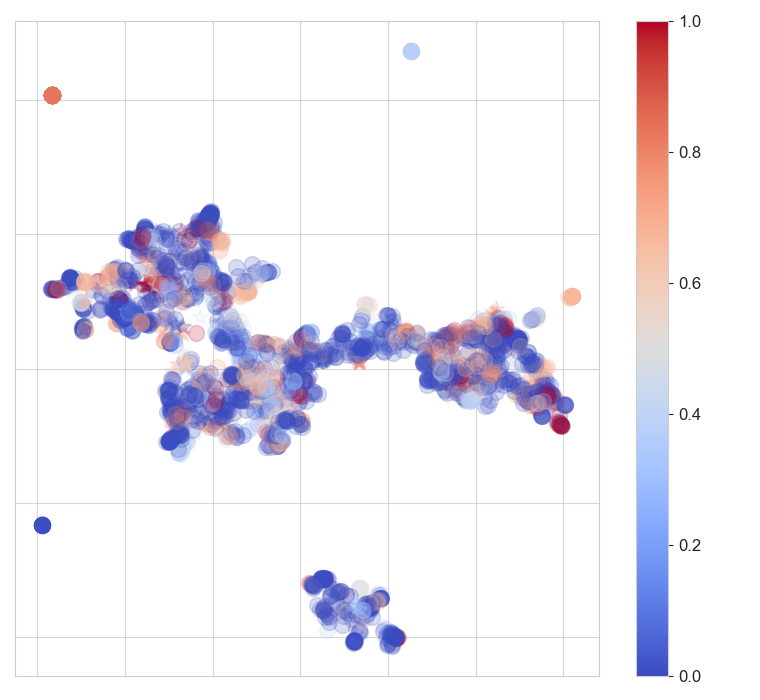}
        \caption{GPT-3.5 on TriviaQA.}\label{subfig:clustering-gpt35}
    \end{subfigure}
    \hfill
    \begin{subfigure}[t]{0.49\columnwidth}
        \centering
        \includegraphics[width=0.98\textwidth]{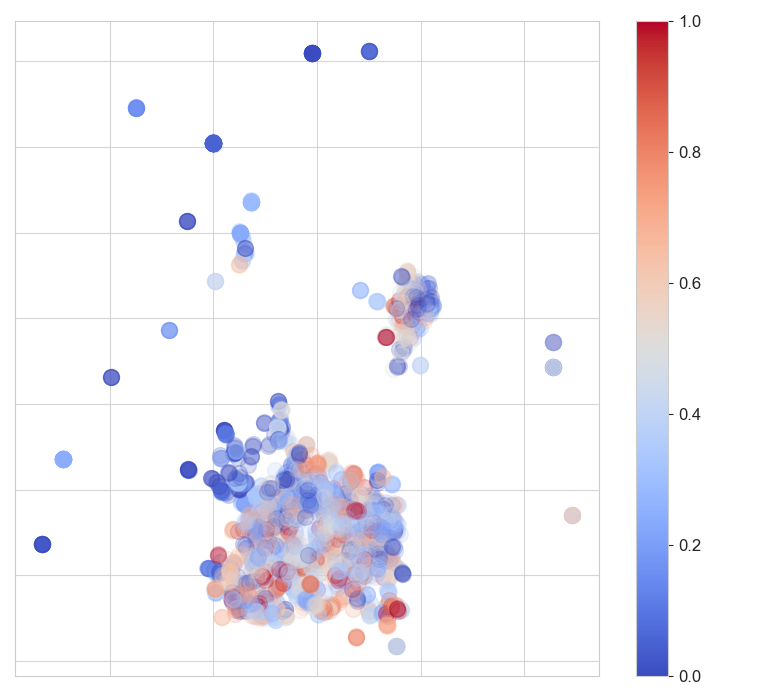}
        \caption{Vicuna v1.5 on CoQA.}\label{subfig:clustering-vicuna-coqa}
    \end{subfigure}
    \hfill
    \begin{subfigure}[t]{0.49\columnwidth}
        \centering
        \includegraphics[width=0.98\textwidth]{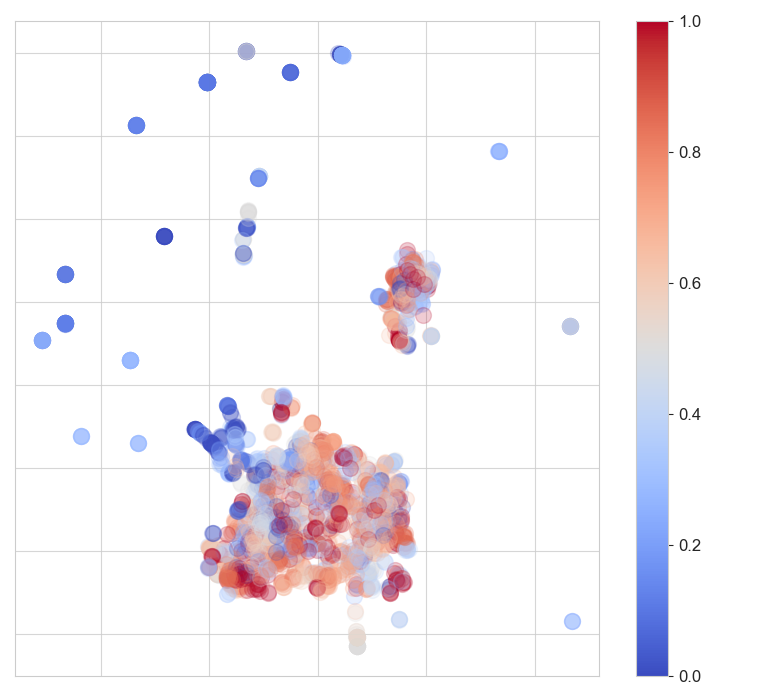}
        \caption{GPT-3.5 on CoQA.}\label{subfig:clustering-gpt35-coqa}
    \end{subfigure}
    \caption{Illustrating questions from TriviaQA along with their assigned confidence targets for the two LLMs, signified through their color from dark blue (0) to dark red (1). To avoid clutter, we subsampled $40 \%$ of the combined datasets to be shown here and used PaCMAP \citep{wang2021understanding} to transform their sentence embeddings into 2D space.}\label{fig:clustering}
\end{figure*}

After clustering and computing the average accuracy per cluster, we obtain a distribution over calibration targets, which we show with density plots in \cref{fig:calibration-targets}.
Since most clusters are of size three, we can see clear modes around $0,\ 0.33,\ 0.66$ and $1$ for Vicuna v1.5 in \cref{subfig:calibration-targets-vicuna}.
For GPT-3.5 in \cref{subfig:calibration-targets-gpt35} these are however less pronounced: We see that targets are often concentrated on $0$ or $1$, respectively.
Similar spikes like in \cref{subfig:calibration-targets-vicuna} are observable for both models on CoQA in \cref{subfig:calibration-targets-coqa-vicuna,subfig:calibration-targets-coqa-gpt-3.5}.
This trend is also visible when plotting the assigned calibration targets per datapoint in \cref{fig:clustering}:
While we can spot more transitionary colors between the blue and red extremes in the manifold for \cref{subfig:clustering-vicuna}, the colors tend more to either of the options \cref{subfig:clustering-gpt35}.
These mode trends continue for CoQA in \cref{subfig:clustering-vicuna-coqa} and \cref{subfig:clustering-gpt35-coqa}.

\subsection{Additional Calibration Results}\label{app:additional-calibration}

\begin{figure*}[htb]
    \centering
    \begin{subfigure}[t]{0.4\columnwidth}
        \centering
        \includegraphics[width=\textwidth]{img/reliability/vicuna-v1.5/test_seq_likelihood.png}
        \caption{Seq. likelihood.}
        \label{subfig:seq-likelihood}
    \end{subfigure}
    \hfill
    \begin{subfigure}[t]{0.4\columnwidth}
        \centering
        \includegraphics[width=\textwidth]{img/reliability/vicuna-v1.5/test_cot_seq_likelihood.png}
        \caption{Seq. like. (CoT).}
        \label{subfig:temperature-scaling}
    \end{subfigure}
    \hfill
    \begin{subfigure}[t]{0.4\columnwidth}
        \centering
        \includegraphics[width=\textwidth]{img/reliability/vicuna-v1.5/test_ts_seq_likelihood.png}
        \caption{Platt scaling.}
        \label{subfig:verbalized-percentage}
    \end{subfigure}
    \hfill
    \begin{subfigure}[t]{0.4\columnwidth}
        \centering
        \includegraphics[width=\textwidth]{img/reliability/vicuna-v1.5/test_ts_cot_seq_likelihood.png}
        \caption{Platt scaling (CoT).}
        \label{subfig:verbalized-percentage}
    \end{subfigure}
    \hfill
    \begin{subfigure}[t]{0.4\columnwidth}
        \centering
        \includegraphics[width=\textwidth]{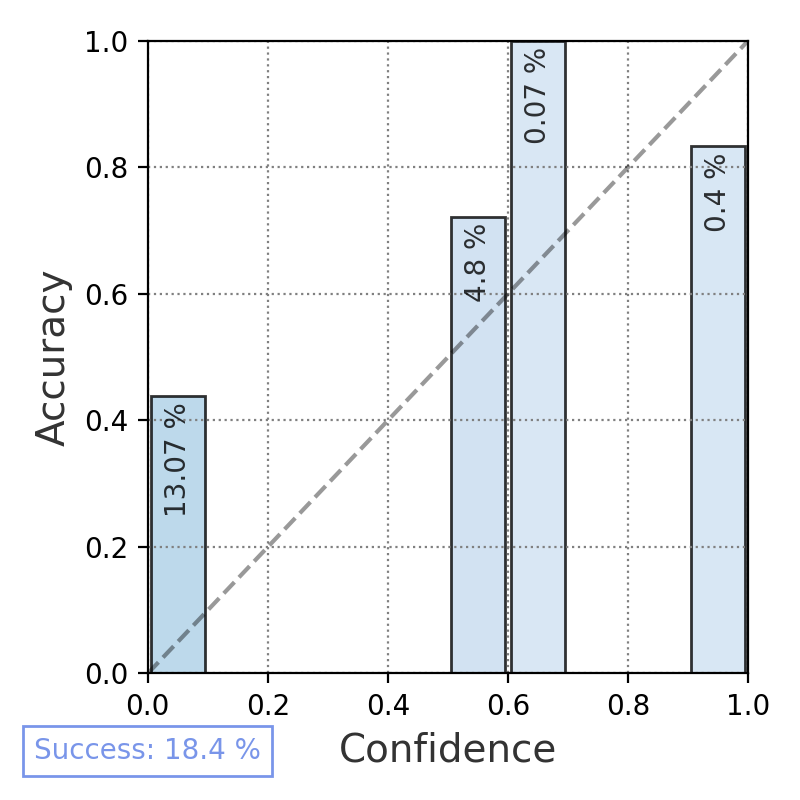}
        \caption{Verbalized Qual.}
        \label{subfig:verbalized-qualitative}
    \end{subfigure} \\
    \begin{subfigure}[t]{0.4\columnwidth}
        \centering
        \includegraphics[width=\textwidth]{img/reliability/vicuna-v1.5/test_verbalized_cot_qual.png}
        \caption{Verb. Qual. (CoT).}
        \label{subfig:seq-likelihood}
    \end{subfigure}
    \hfill
    \begin{subfigure}[t]{0.4\columnwidth}
        \centering
        \includegraphics[width=\textwidth]{img/reliability/vicuna-v1.5/test_verbalized_quant.png}
        \caption{Verbalized Quant.}
        \label{subfig:temperature-scaling}
    \end{subfigure}
    \hfill
    \begin{subfigure}[t]{0.4\columnwidth}
        \centering
        \includegraphics[width=\textwidth]{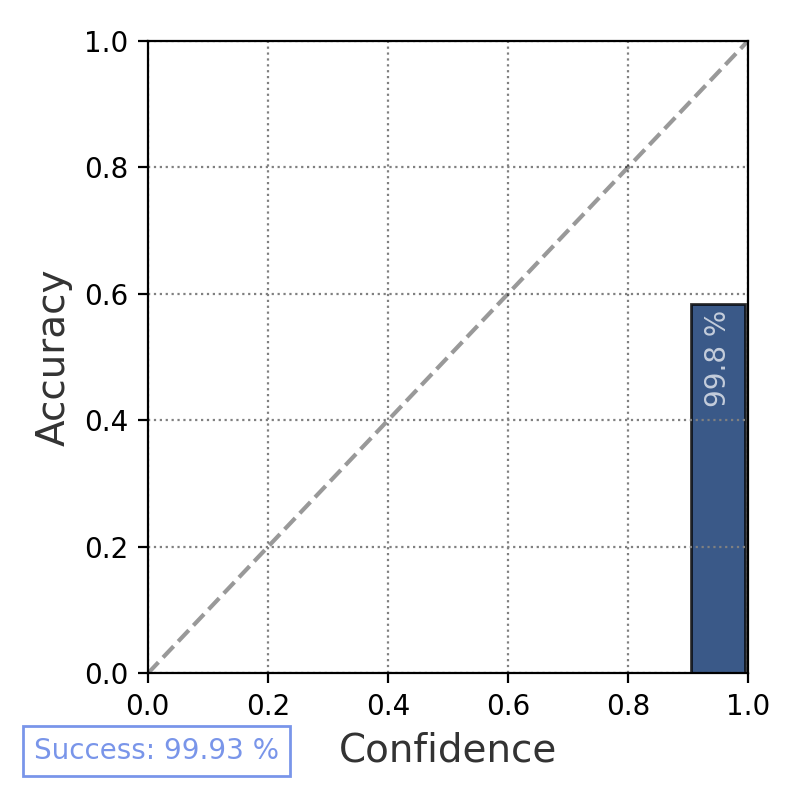}
        \caption{Verb. Quant. (CoT).}
        \label{subfig:verbalized-percentage}
    \end{subfigure}
    \hfill
    \begin{subfigure}[t]{0.4\columnwidth}
        \centering
        \includegraphics[width=\textwidth]{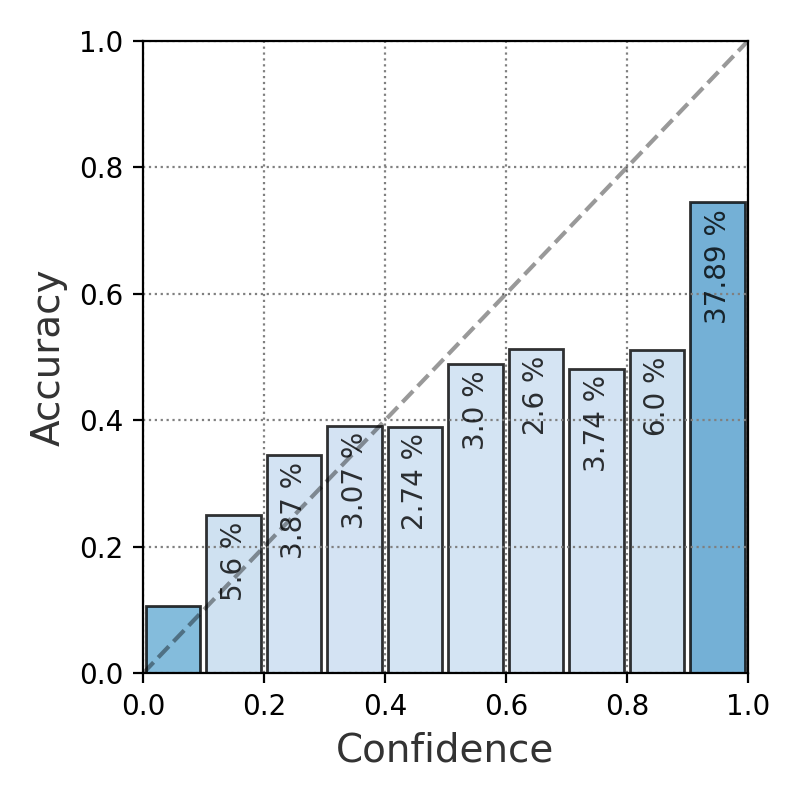}
        \caption{Auxiliary (binary).}
        \label{subfig:verbalized-qualitative}
    \end{subfigure}
    \hfill
    \begin{subfigure}[t]{0.4\columnwidth}
        \centering
        \includegraphics[width=\textwidth]{img/reliability/vicuna-v1.5/16-02-2024_test_answer_question.png}
        \caption{Aux. (clustering).}
        \label{subfig:verbalized-qualitative}
    \end{subfigure}
    \caption{Reliability diagrams for our different methods using $10$ bins each for Vicuna v1.5 7B on TriviaQA. The color as well as the percentage number within each bar indicate the proportion of total points contained in each bin.}\label{fig:reliabiliy-diagrams-vicuna-trivia-qa-full}
\end{figure*}

\begin{figure*}[htb]
    \centering
    \begin{subfigure}[t]{0.4\columnwidth}
        \centering
        \includegraphics[width=\textwidth]{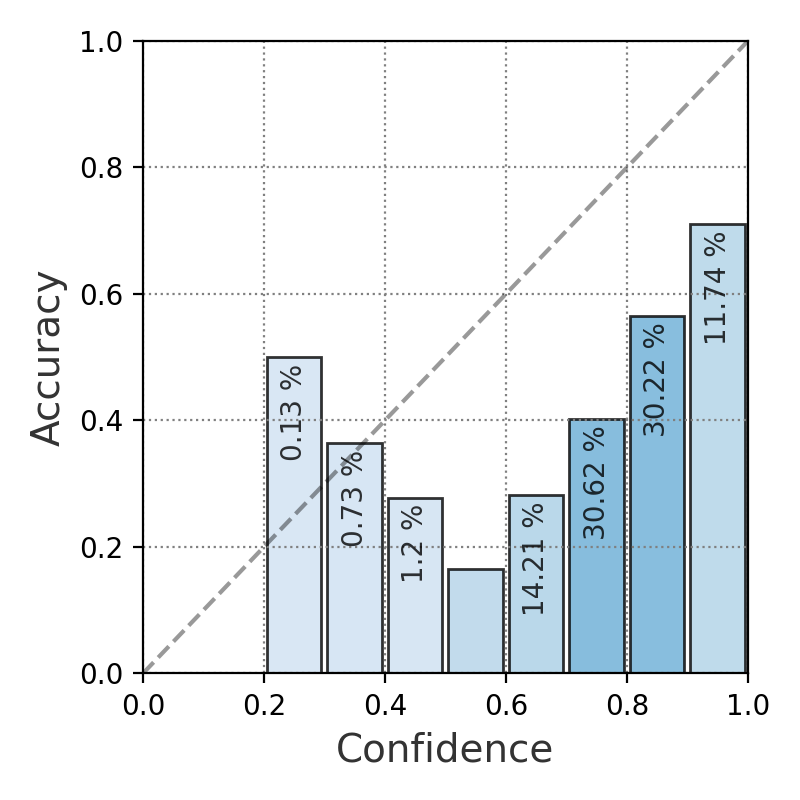}
        \caption{Seq. likelihood.}
        \label{subfig:seq-likelihood}
    \end{subfigure}
    \hfill
    \begin{subfigure}[t]{0.4\columnwidth}
        \centering
        \includegraphics[width=\textwidth]{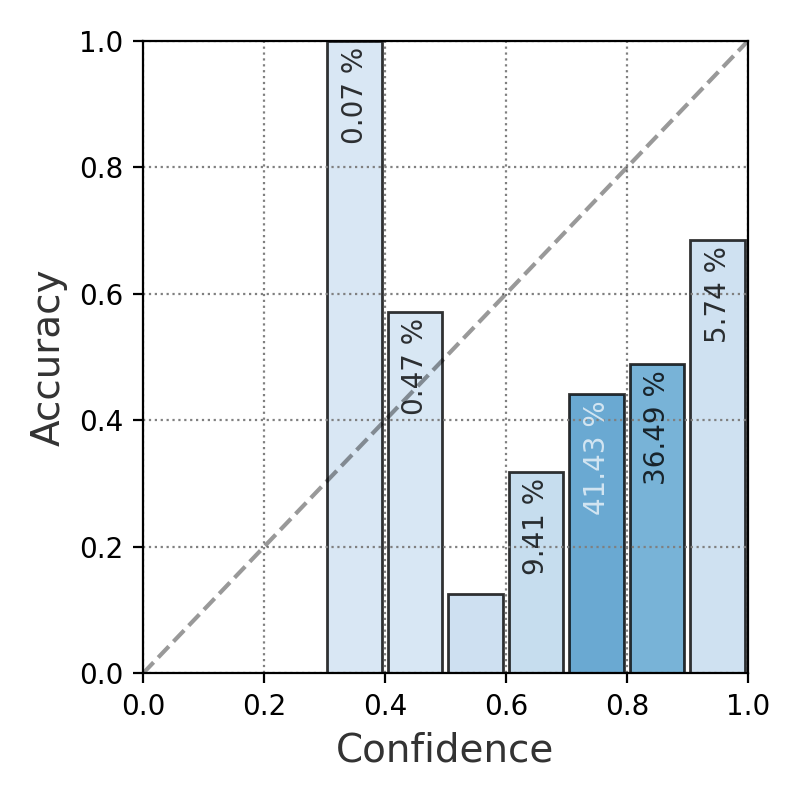}
        \caption{Seq. like. (CoT).}
        \label{subfig:temperature-scaling}
    \end{subfigure}
    \hfill
    \begin{subfigure}[t]{0.4\columnwidth}
        \centering
        \includegraphics[width=\textwidth]{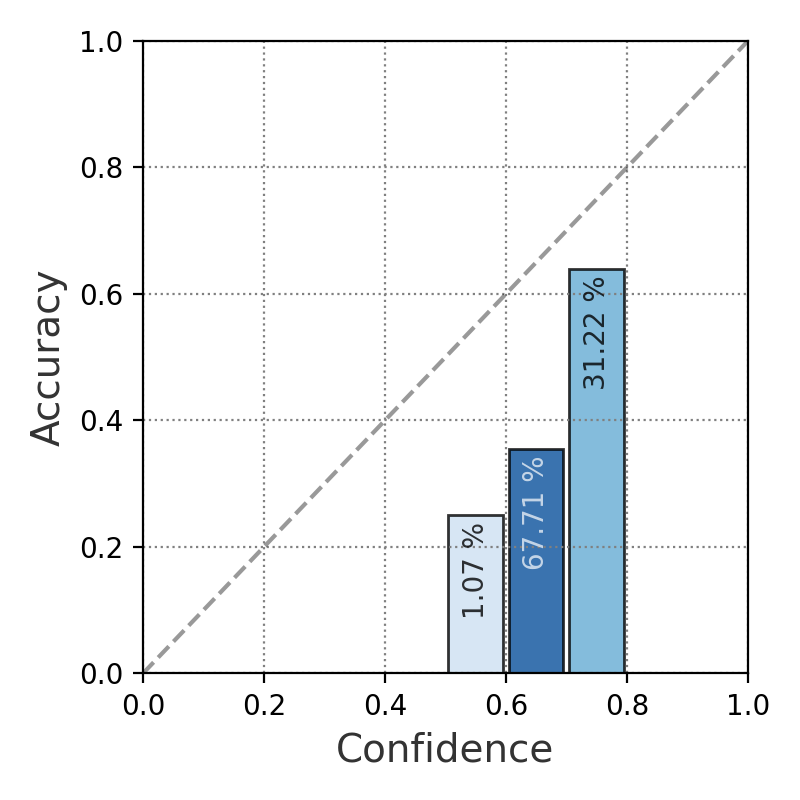}
        \caption{Platt scaling.}
        \label{subfig:verbalized-percentage}
    \end{subfigure}
    \hfill
    \begin{subfigure}[t]{0.4\columnwidth}
        \centering
        \includegraphics[width=\textwidth]{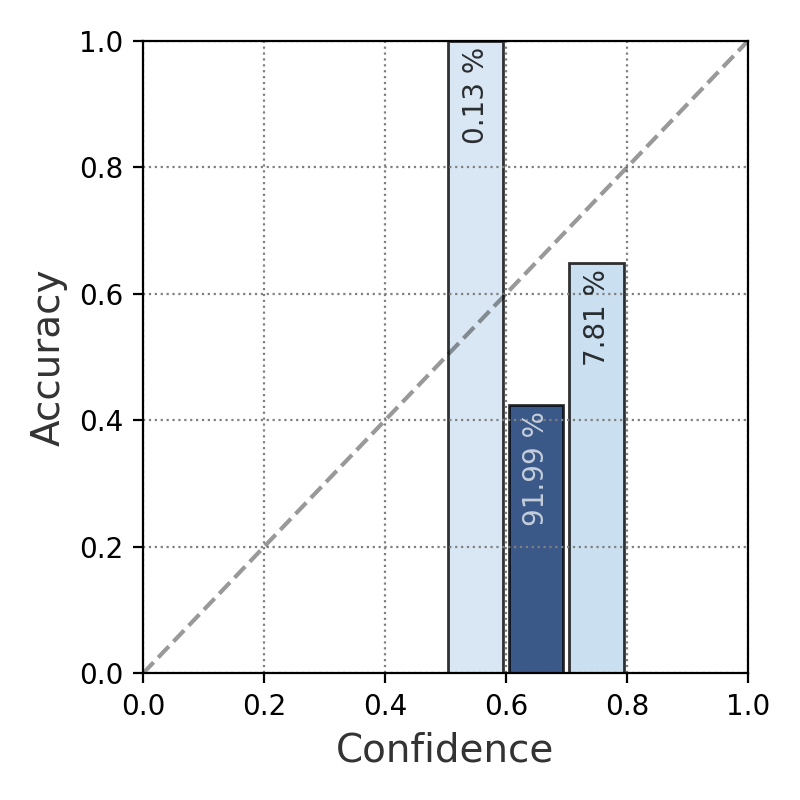}
        \caption{Platt scaling (CoT).}
        \label{subfig:verbalized-percentage}
    \end{subfigure}
    \hfill
    \begin{subfigure}[t]{0.4\columnwidth}
        \centering
        \includegraphics[width=\textwidth]{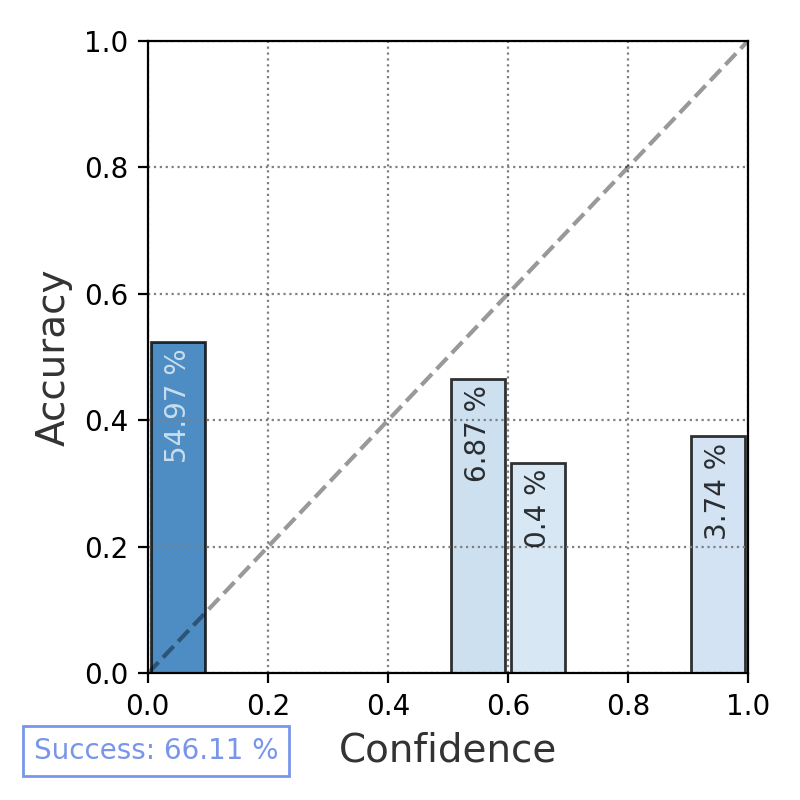}
        \caption{Verbalized Qual.}
        \label{subfig:verbalized-qualitative}
    \end{subfigure} \\
    \begin{subfigure}[t]{0.4\columnwidth}
        \centering
        \includegraphics[width=\textwidth]{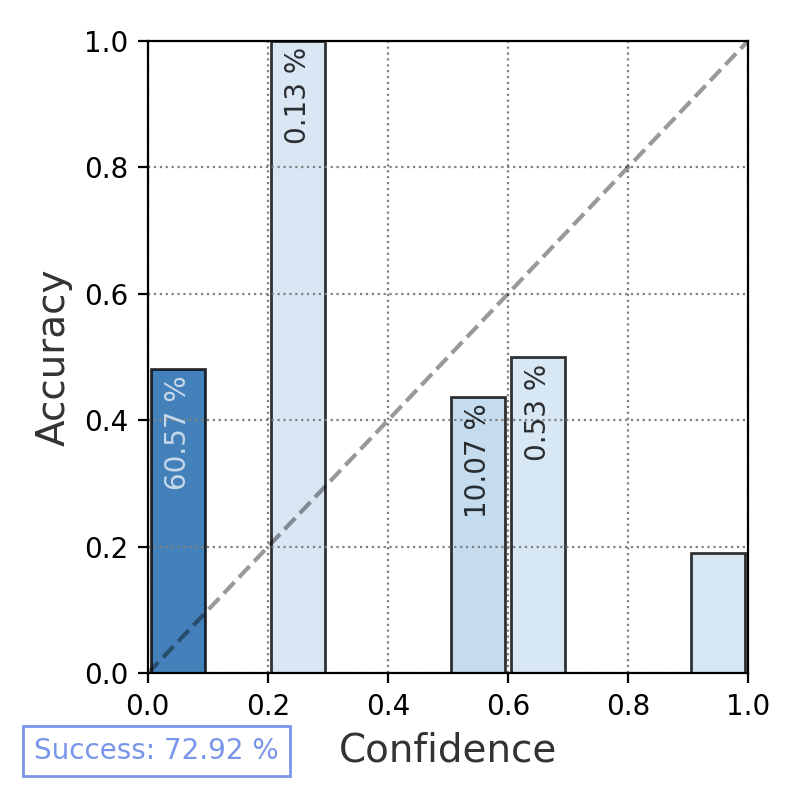}
        \caption{Verb. Qual. (CoT).}
        \label{subfig:seq-likelihood}
    \end{subfigure}
    \hfill
    \begin{subfigure}[t]{0.4\columnwidth}
        \centering
        \includegraphics[width=\textwidth]{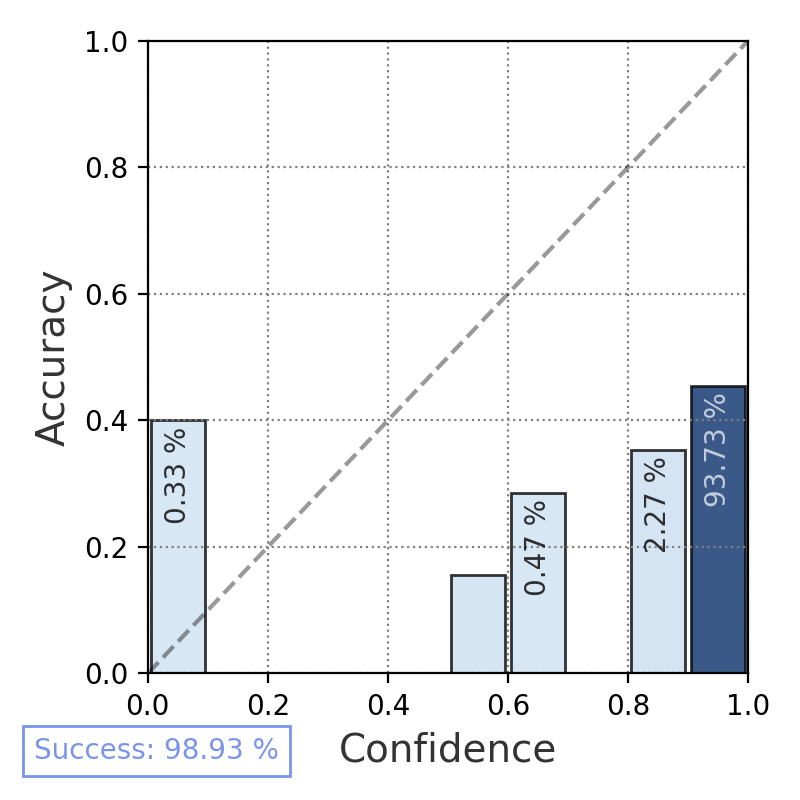}
        \caption{Verbalized $\%$.}
        \label{subfig:temperature-scaling}
    \end{subfigure}
    \hfill
    \begin{subfigure}[t]{0.4\columnwidth}
        \centering
        \includegraphics[width=\textwidth]{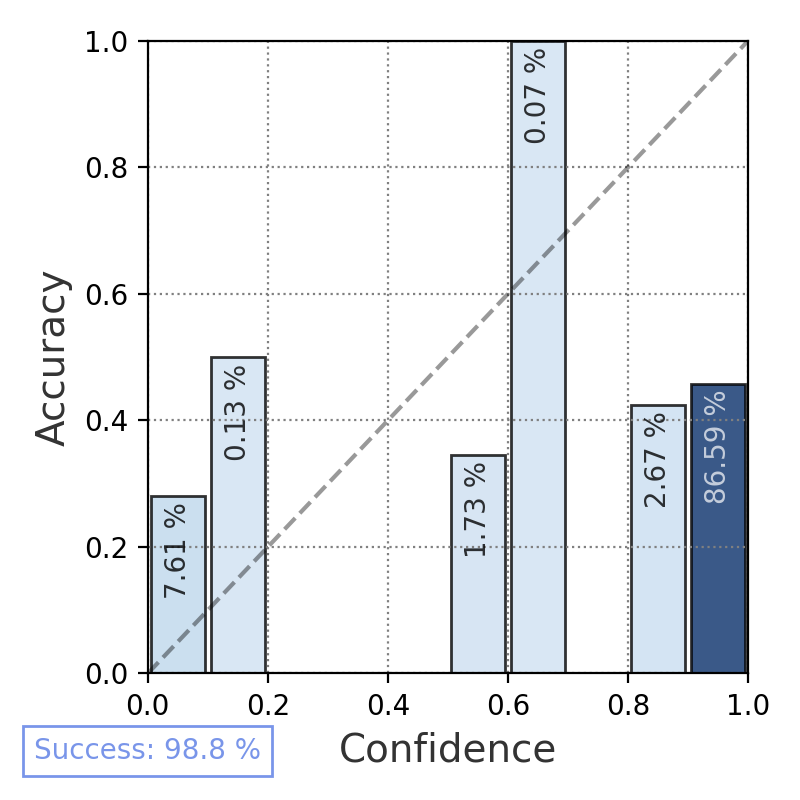}
        \caption{Verb. $\%$ (CoT).}
        \label{subfig:verbalized-percentage}
    \end{subfigure}
    \hfill
    \begin{subfigure}[t]{0.4\columnwidth}
        \centering
        \includegraphics[width=\textwidth]{img/reliability/vicuna-v1.5-coqa/24-02-2024_test_binary_answer_question.png}
        \caption{Auxiliary (binary).}
        \label{subfig:verbalized-qualitative}
    \end{subfigure}
    \hfill
    \begin{subfigure}[t]{0.4\columnwidth}
        \centering
        \includegraphics[width=\textwidth]{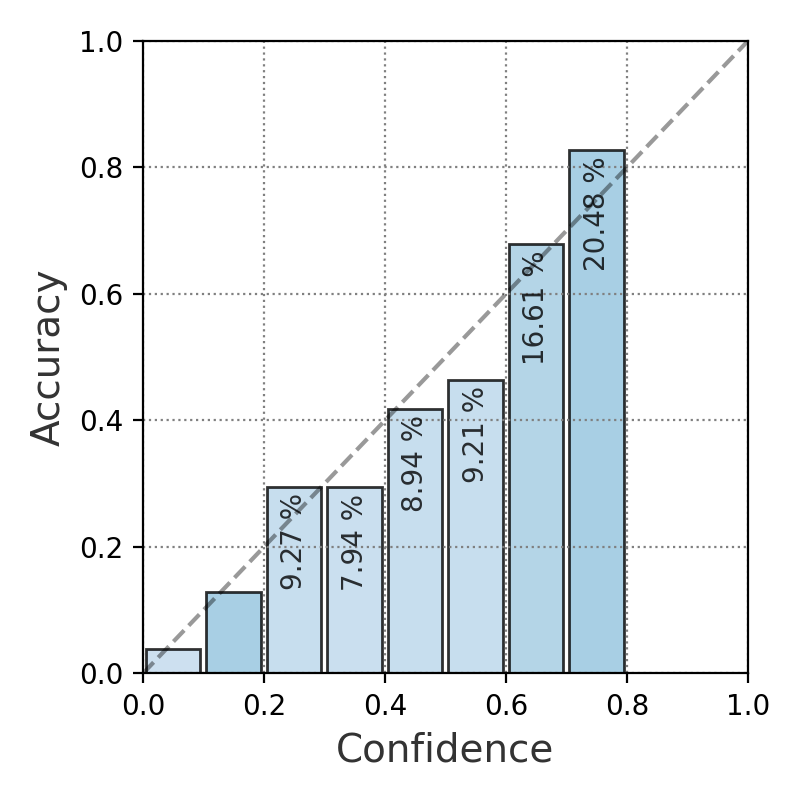}
        \caption{Aux. (clustering).}
        \label{subfig:verbalized-qualitative}
    \end{subfigure}
    \caption{Reliability diagrams for our different methods using $10$ bins each for Vicuna v1.5 7B on CoQA. The color as well as the percentage number within each bar indicate the proportion of total points contained in each bin.}\label{fig:reliabiliy-diagrams-vicuna-coqa-full}
\end{figure*}

\begin{figure*}[htb]
    \centering
    \begin{subfigure}[t]{0.4\columnwidth}
        \centering
        \includegraphics[width=\textwidth]{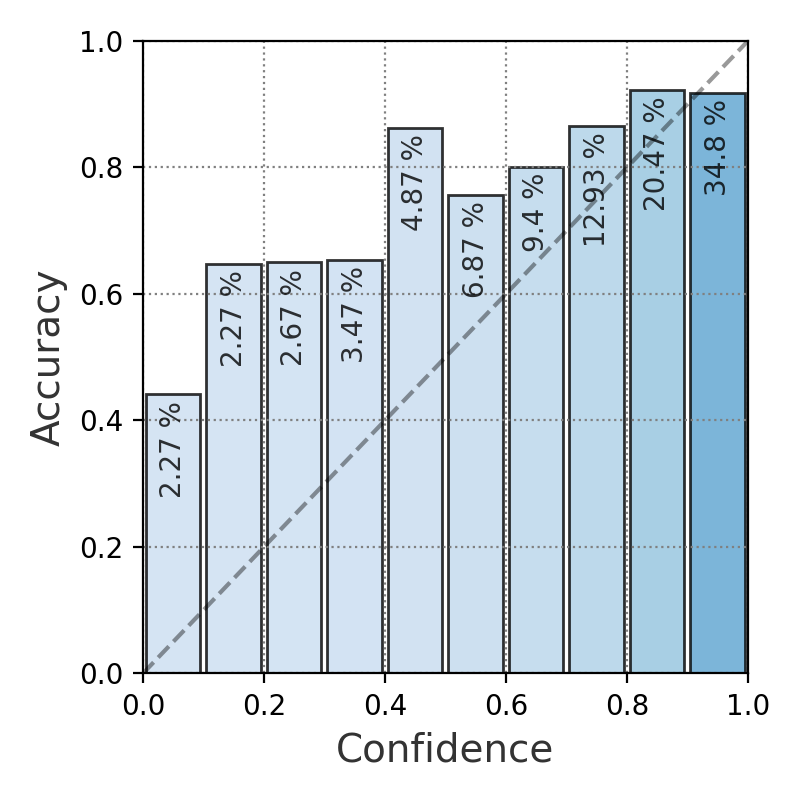}
        \caption{Seq. likelihood.}
        \label{subfig:seq-likelihood}
    \end{subfigure}
    \hfill
    \begin{subfigure}[t]{0.4\columnwidth}
        \centering
        \includegraphics[width=\textwidth]{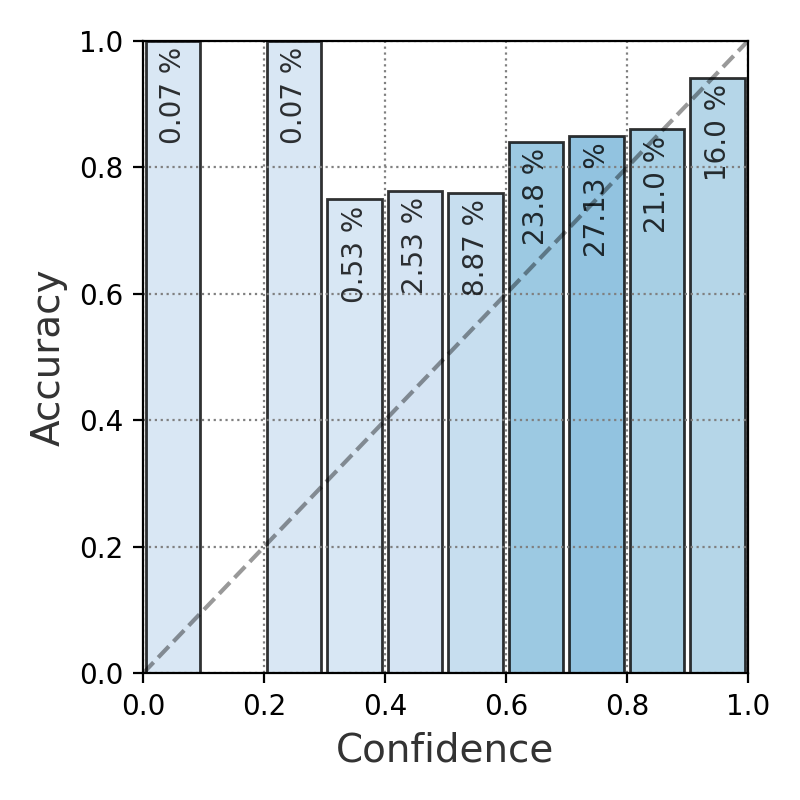}
        \caption{Seq. like. (CoT).}
        \label{subfig:temperature-scaling}
    \end{subfigure}
    \hfill
    \begin{subfigure}[t]{0.4\columnwidth}
        \centering
        \includegraphics[width=\textwidth]{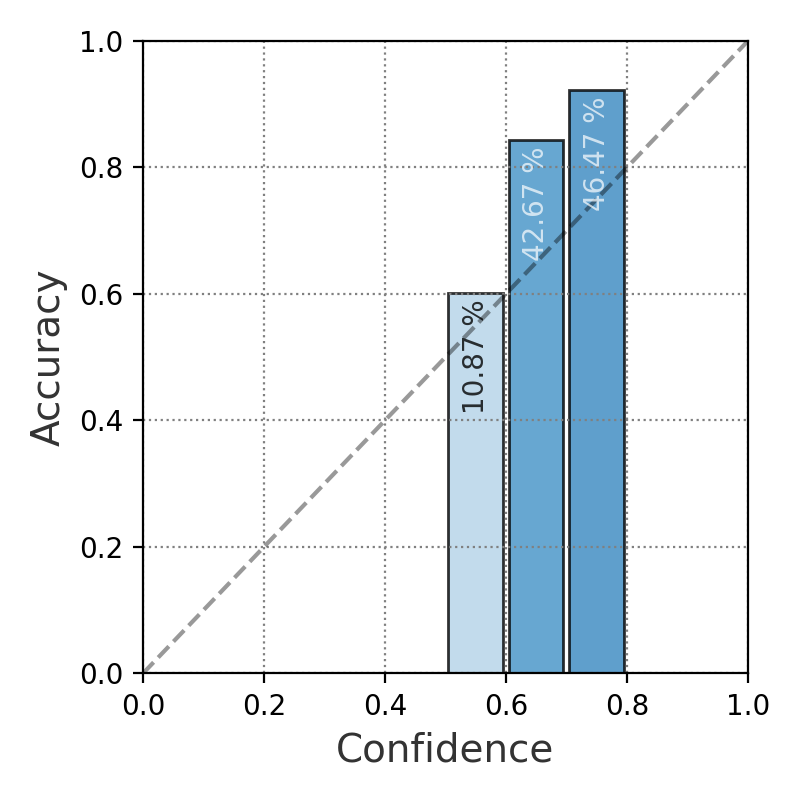}
        \caption{Platt scaling.}
        \label{subfig:verbalized-percentage}
    \end{subfigure}
    \hfill
    \begin{subfigure}[t]{0.4\columnwidth}
        \centering
        \includegraphics[width=\textwidth]{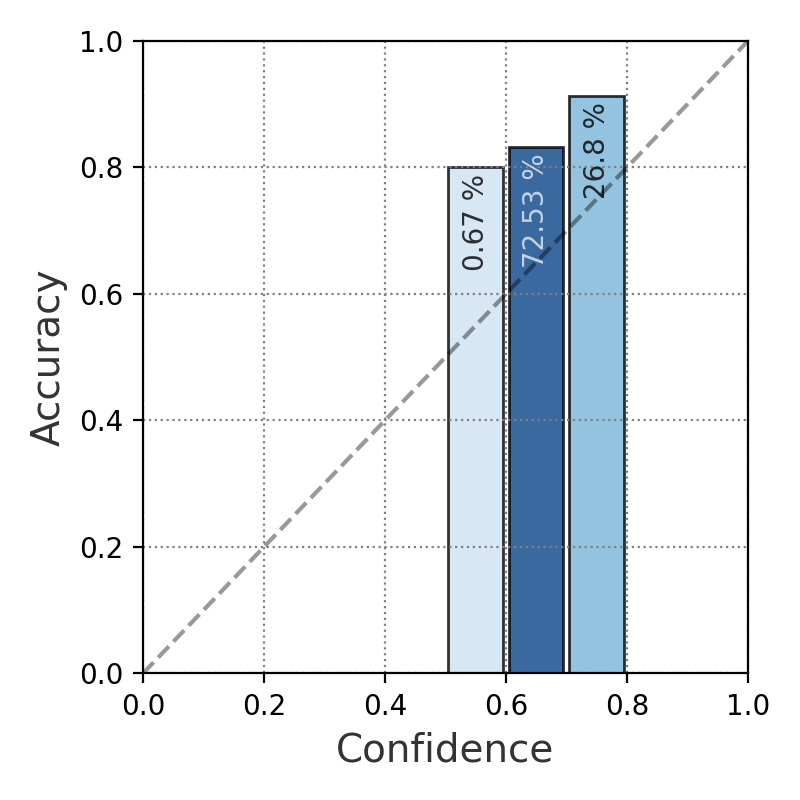}
        \caption{Platt scaling (CoT).}
        \label{subfig:verbalized-percentage}
    \end{subfigure}
    \hfill
    \begin{subfigure}[t]{0.4\columnwidth}
        \centering
        \includegraphics[width=\textwidth]{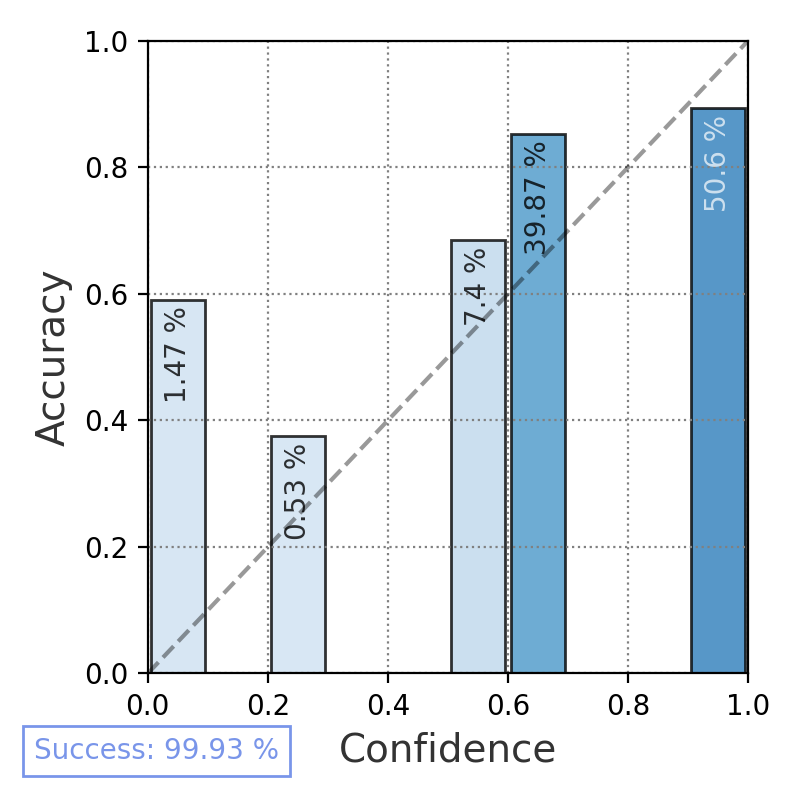}
        \caption{Verbalized Qual.}
        \label{subfig:verbalized-qualitative}
    \end{subfigure} \\
    \begin{subfigure}[t]{0.4\columnwidth}
        \centering
        \includegraphics[width=\textwidth]{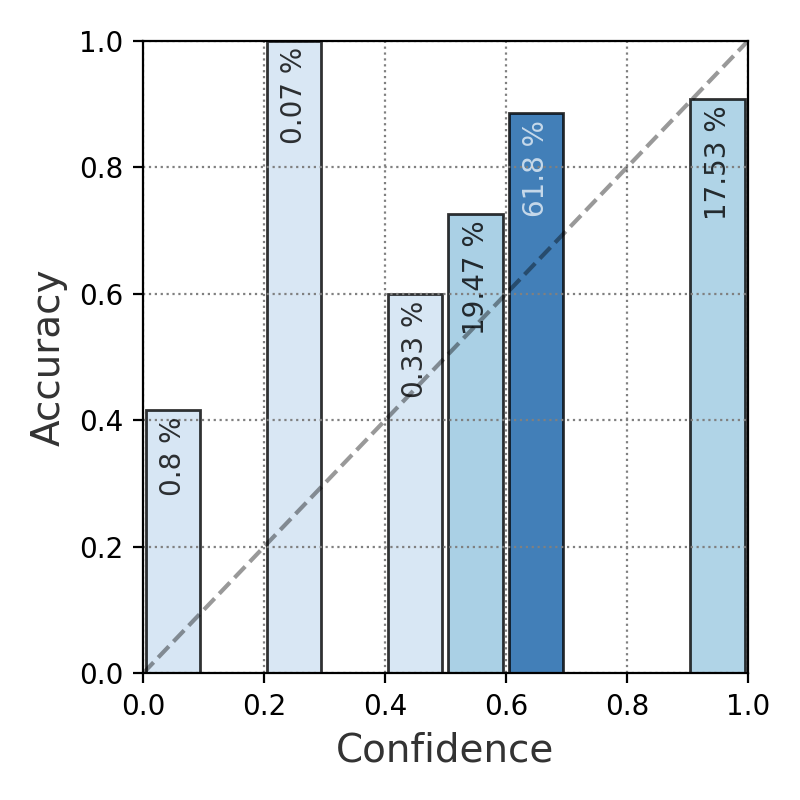}
        \caption{Verb. Qual. (CoT).}
        \label{subfig:seq-likelihood}
    \end{subfigure}
    \hfill
    \begin{subfigure}[t]{0.4\columnwidth}
        \centering
        \includegraphics[width=\textwidth]{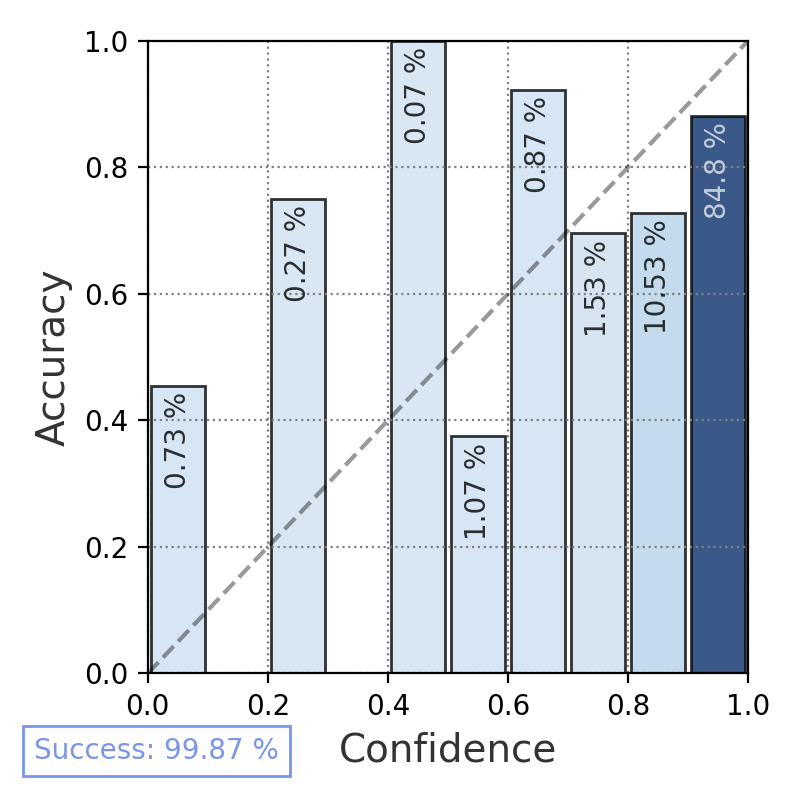}
        \caption{Verbalized $\%$.}
        \label{subfig:temperature-scaling}
    \end{subfigure}
    \hfill
    \begin{subfigure}[t]{0.4\columnwidth}
        \centering
        \includegraphics[width=\textwidth]{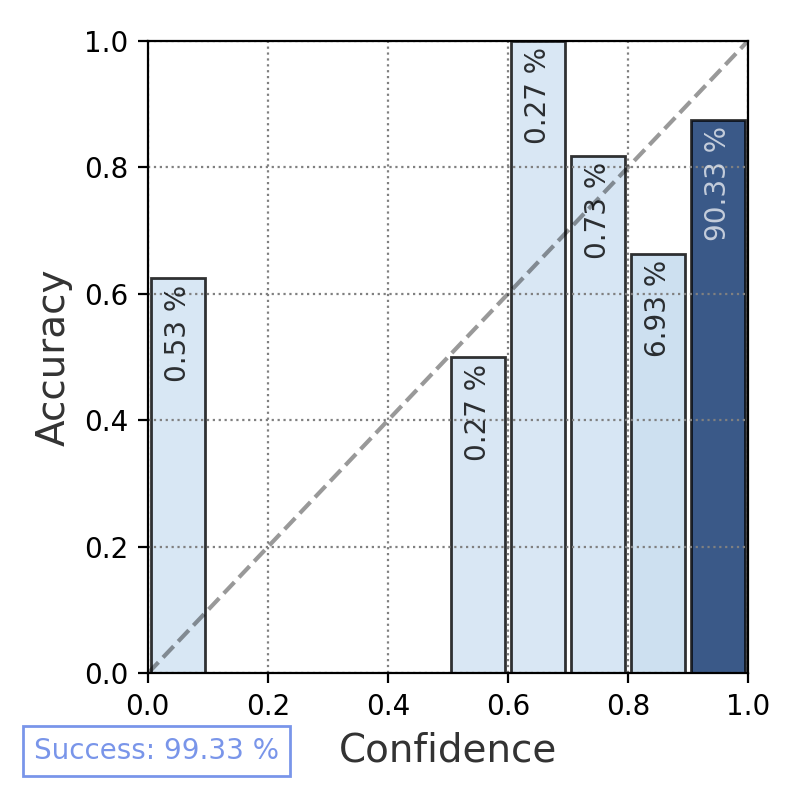}
        \caption{Verb. $\%$ (CoT).}
        \label{subfig:verbalized-percentage}
    \end{subfigure}
    \hfill
    \begin{subfigure}[t]{0.4\columnwidth}
        \centering
        \includegraphics[width=\textwidth]{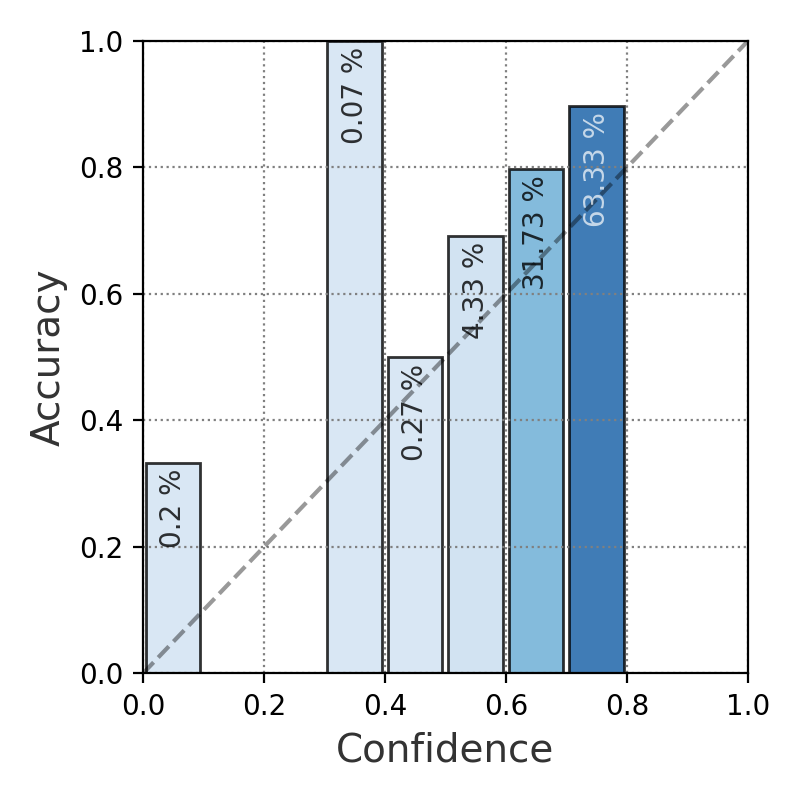}
        \caption{Auxiliary (binary).}
        \label{subfig:verbalized-qualitative}
    \end{subfigure}
    \hfill
    \begin{subfigure}[t]{0.4\columnwidth}
        \centering
        \includegraphics[width=\textwidth]{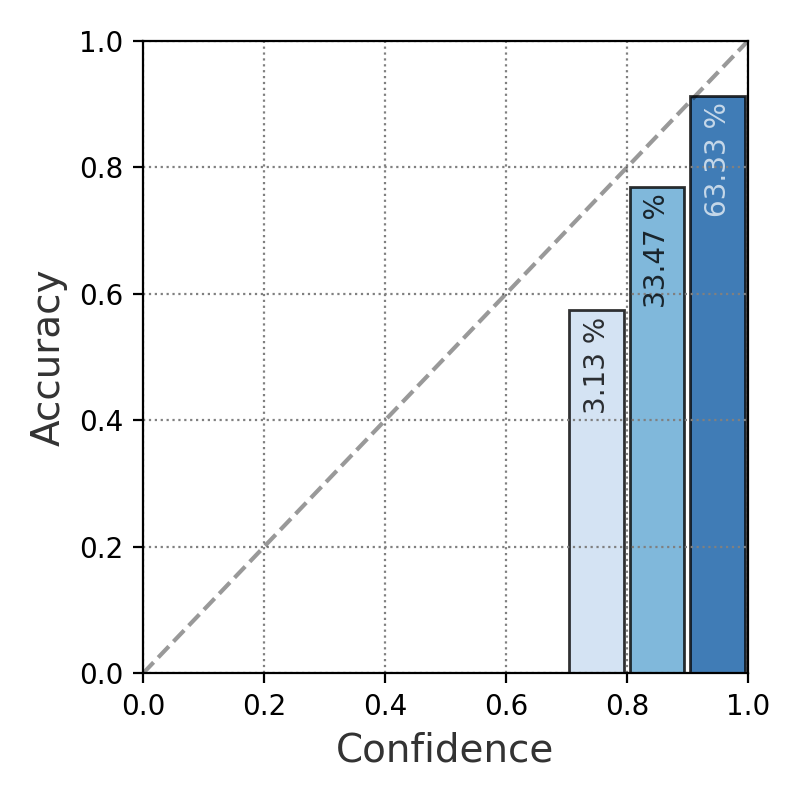}
        \caption{Aux. (clustering).}
        \label{subfig:verbalized-qualitative}
    \end{subfigure}
    \caption{Reliability diagrams for our different methods using $10$ bins each for GPT-3.5 on TriviaQA. The color as well as the percentage number within each bar indicate the proportion of total points contained in each bin.}\label{fig:reliabiliy-diagrams-gpt35-trivia-qa-full}
\end{figure*}

\begin{figure*}[htb]
    \centering
    \begin{subfigure}[t]{0.4\columnwidth}
        \centering
        \includegraphics[width=\textwidth]{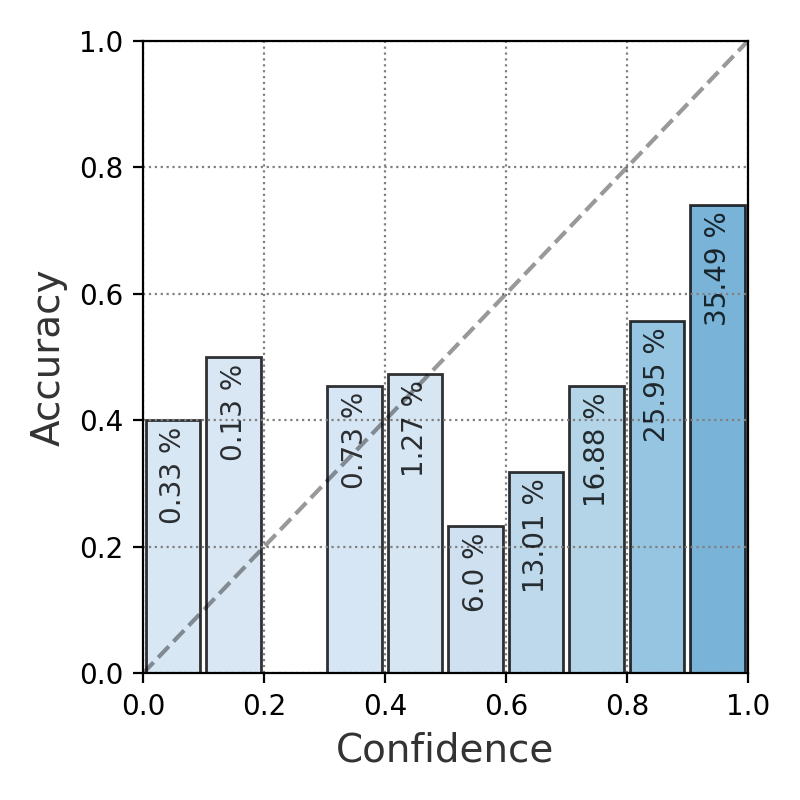}
        \caption{Seq. likelihood.}
        \label{subfig:seq-likelihood}
    \end{subfigure}
    \hfill
    \begin{subfigure}[t]{0.4\columnwidth}
        \centering
        \includegraphics[width=\textwidth]{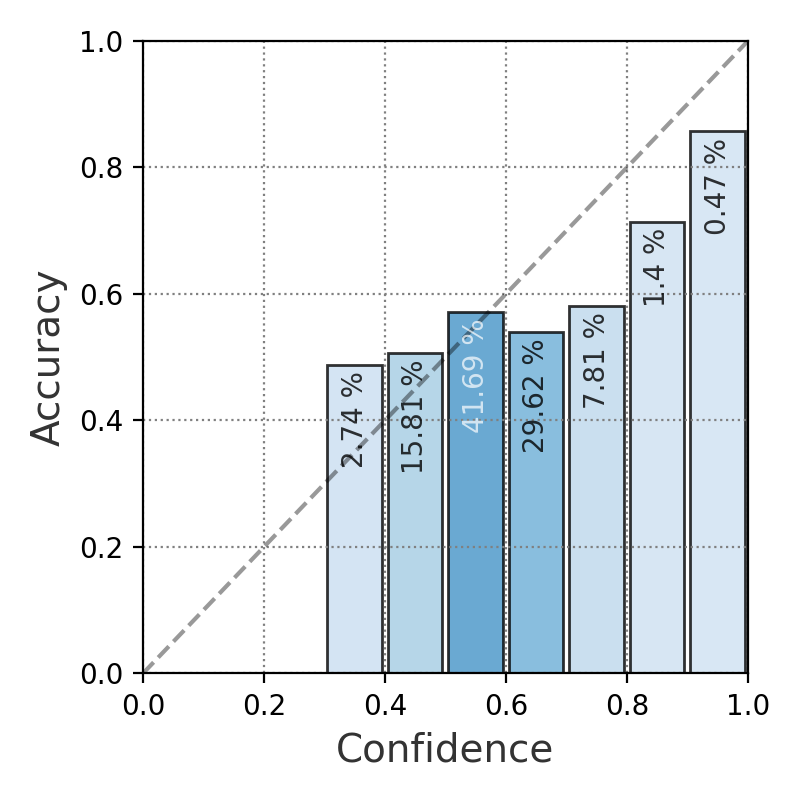}
        \caption{Seq. like. (CoT).}
        \label{subfig:temperature-scaling}
    \end{subfigure}
    \hfill
    \begin{subfigure}[t]{0.4\columnwidth}
        \centering
        \includegraphics[width=\textwidth]{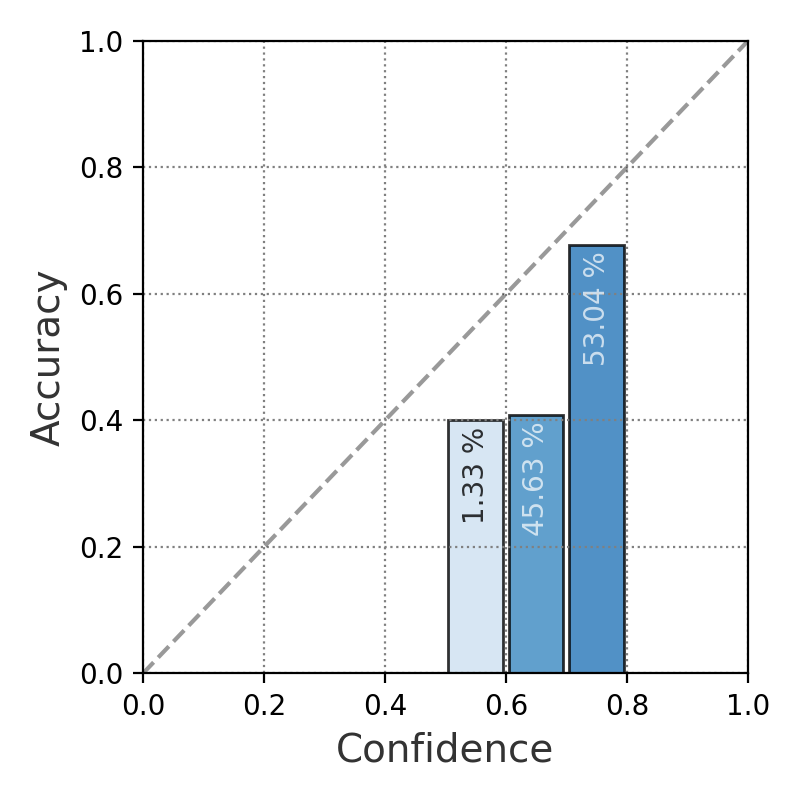}
        \caption{Platt scaling.}
        \label{subfig:verbalized-percentage}
    \end{subfigure}
    \hfill
    \begin{subfigure}[t]{0.4\columnwidth}
        \centering
        \includegraphics[width=\textwidth]{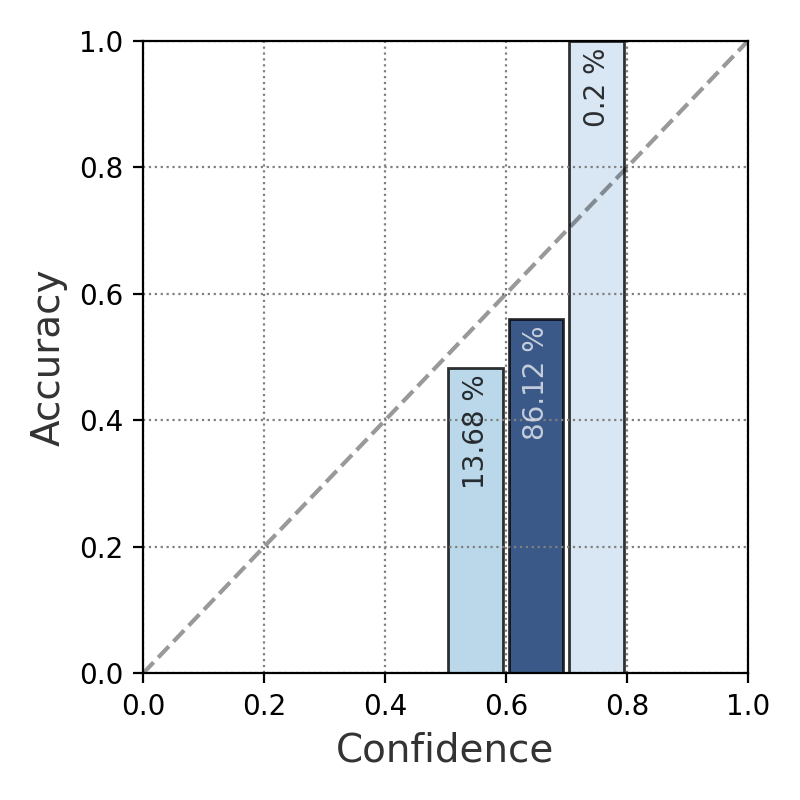}
        \caption{Platt scaling (CoT).}
        \label{subfig:verbalized-percentage}
    \end{subfigure}
    \hfill
    \begin{subfigure}[t]{0.4\columnwidth}
        \centering
        \includegraphics[width=\textwidth]{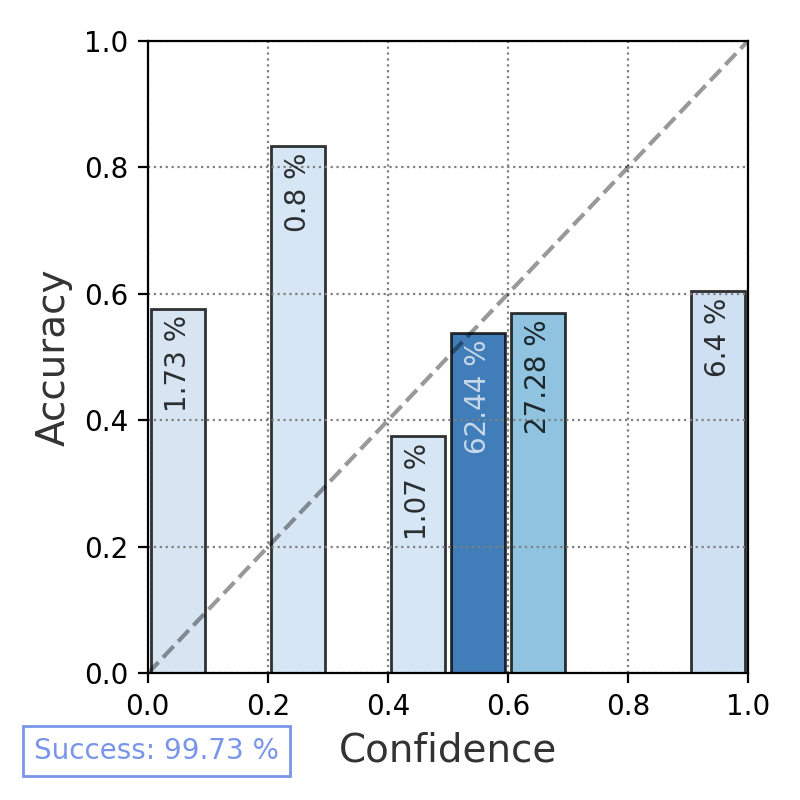}
        \caption{Verbalized Qual.}
        \label{subfig:verbalized-qualitative}
    \end{subfigure} \\
    \begin{subfigure}[t]{0.4\columnwidth}
        \centering
        \includegraphics[width=\textwidth]{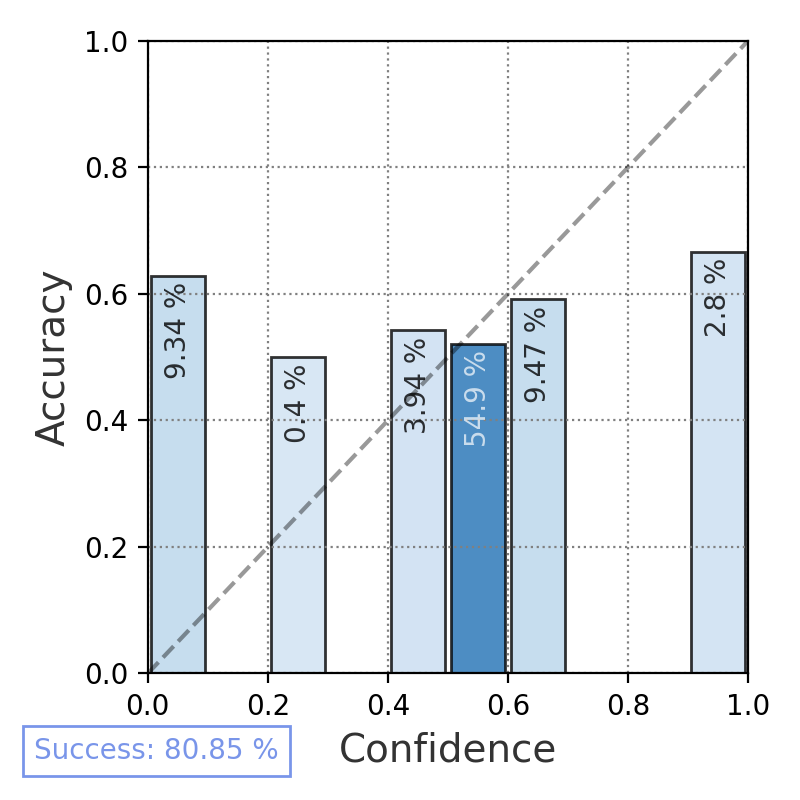}
        \caption{Verb. Qual. (CoT).}
        \label{subfig:seq-likelihood}
    \end{subfigure}
    \hfill
    \begin{subfigure}[t]{0.4\columnwidth}
        \centering
        \includegraphics[width=\textwidth]{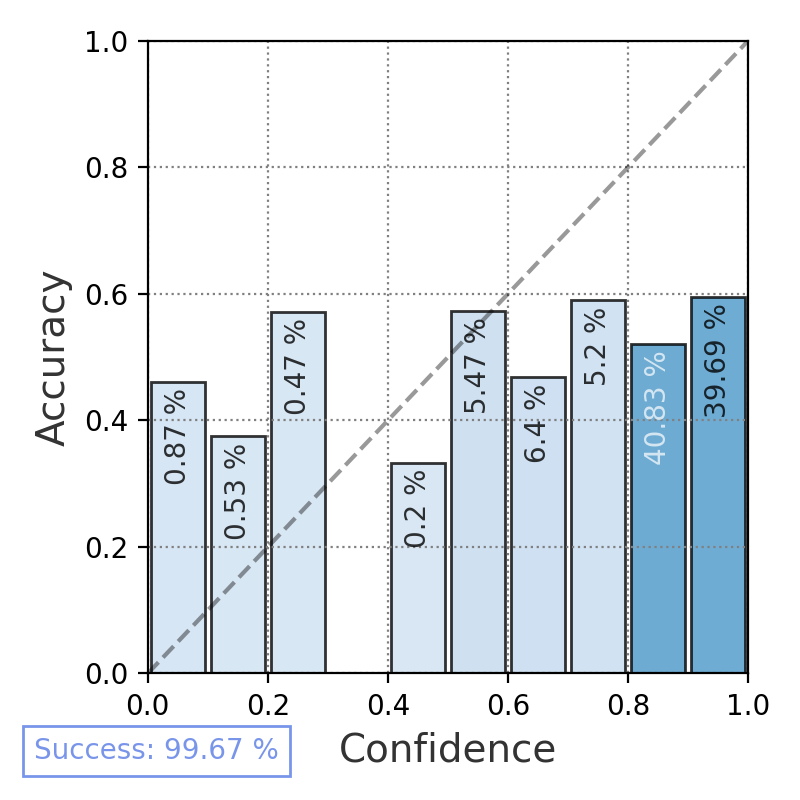}
        \caption{Verbalized $\%$.}
        \label{subfig:temperature-scaling}
    \end{subfigure}
    \hfill
    \begin{subfigure}[t]{0.4\columnwidth}
        \centering
        \includegraphics[width=\textwidth]{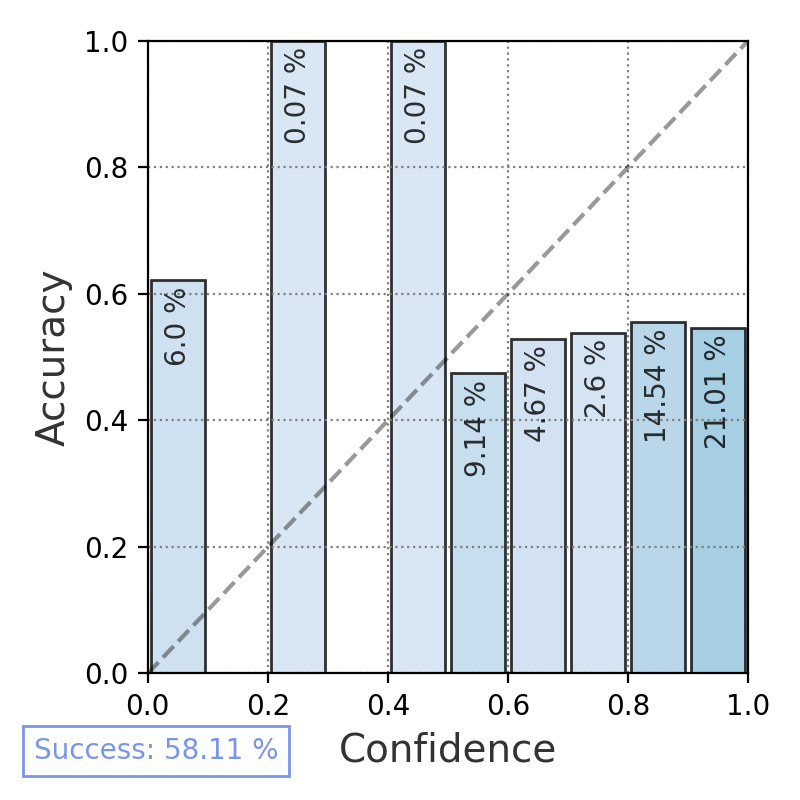}
        \caption{Verb. $\%$ (CoT).}
        \label{subfig:verbalized-percentage}
    \end{subfigure}
    \hfill
    \begin{subfigure}[t]{0.4\columnwidth}
        \centering
        \includegraphics[width=\textwidth]{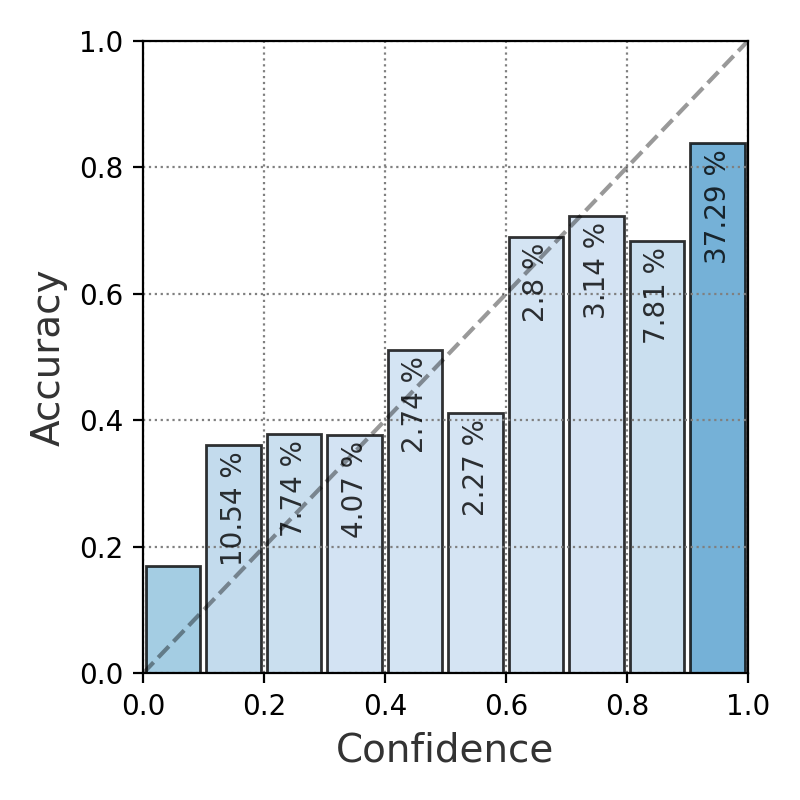}
        \caption{Auxiliary (binary).}
        \label{subfig:verbalized-qualitative}
    \end{subfigure}
    \hfill
    \begin{subfigure}[t]{0.4\columnwidth}
        \centering
        \includegraphics[width=\textwidth]{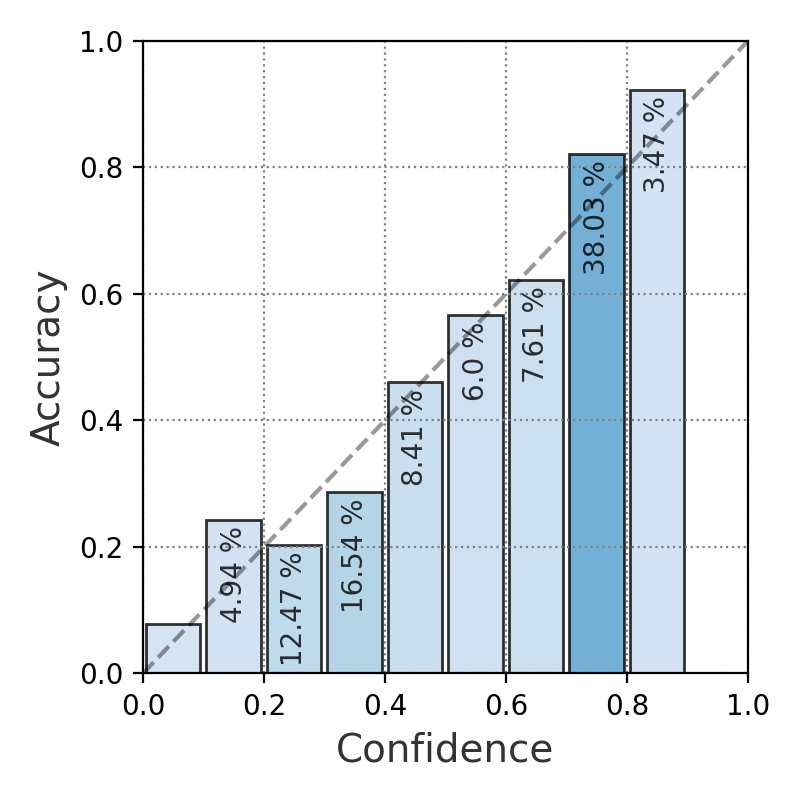}
        \caption{Aux. (clustering).}
        \label{subfig:verbalized-qualitative}
    \end{subfigure}
    \caption{Reliability diagrams for our different methods using $10$ bins each for GPT-3.5 on CoQA. The color as well as the percentage number within each bar indicate the proportion of total points contained in each bin.}\label{fig:reliabiliy-diagrams-gpt35-coqa-full}
\end{figure*}

\paragraph{Additional reliability plots.} We show the all available reliability diagrams for Vicuna-v1.5 for TriviaQA in \cref{fig:reliabiliy-diagrams-vicuna-trivia-qa} and CoQA in \cref{fig:reliabiliy-diagrams-vicuna-coqa-full}, as well as the corresponding plots for GPT-3.5 in \cref{fig:reliabiliy-diagrams-gpt35-trivia-qa-full,fig:reliabiliy-diagrams-gpt35-coqa-full}.
We can summarize some general trends:
Sequence likelihood \emph{can} be well-calibrated already, but this fact depends strongly on the dataset in question.
And while our version of Platt scaling can improve results, it also narrows the range of confidence values to a narrow window.
Verbalized uncertainty in both of variants also is not able to produce a wide variety of responses, even though this effect is slightly less pronounced for GPT-3.5.
This could be explained by previous observations by \citet{zhou2023navigating} that certain mentions of percentage values are over-represented in model training data.
The auxiliary model is able to predict a wider array of confidence values in all settings, with the clustering variant achieving better calibration overall.

\begin{table*}
    \renewcommand{\arraystretch}{1.5}
    \centering
    \resizebox{1.975\columnwidth}{!}{
    \begin{tabular}{@{}lcccccccccccccc@{}}
        \toprule
         & \multicolumn{4}{c}{Auxiliary Model Input}  & \multicolumn{5}{c}{TriviaQA} & \multicolumn{5}{c}{CoQA} \\
         \cmidrule(lr){2-5} \cmidrule(lr){6-9} \cmidrule(lr){10-13}
        & Quest. & Ans. & CoT & Verb. & Brier$\downarrow$ & ECE$\downarrow$  & smECE$\downarrow$ & AUROC$\uparrow$ & Brier$\downarrow$ & ECE$\downarrow$  & smECE$\downarrow$ & AUROC$\uparrow$  \\
        \toprule
        \multirow{7}{*}{\rotatebox{90}{Vicuna v1.5 (white-box)}} 
        & \gcheck & \rcross & \rcross & \rcross & $.21{\scriptstyle\ \pm.00}$& $\mathbf{.07}{\scriptstyle\ \pm.01}$& $\underline{\mathbf{.06}}{\scriptstyle\ \pm.01}$& $.74{\scriptstyle\ \pm.01}$ & $.22{\scriptstyle\ \pm.00}$& $\underline{\mathbf{.03}}{\scriptstyle\ \pm.01}$& $\mathbf{.03}{\scriptstyle\ \pm.00}$& $.70{\scriptstyle\ \pm.01}$ \\
        & \gcheck & \gcheck & \rcross & \rcross & $\mathbf{.18}{\scriptstyle\ \pm.00}$& $.09{\scriptstyle\ \pm.01}$& $.09{\scriptstyle\ \pm.01}$& $\underline{\mathbf{.83}}{\scriptstyle\ \pm.01}$ & $\underline{\mathbf{.18}}{\scriptstyle\ \pm.00}$& $.04{\scriptstyle\ \pm.01}$& $.04{\scriptstyle\ \pm.01}$& $\underline{\mathbf{.82}}{\scriptstyle\ \pm.01}$ \\
        & \gcheck & \gcheck & \rcross & {\color{green!55!black}\textbf{Qual.}} & $\mathbf{.18}{\scriptstyle\ \pm.00}$& $.08{\scriptstyle\ \pm.01}$& $.08{\scriptstyle\ \pm.01}$& $.82{\scriptstyle\ \pm.01}$ & $.19{\scriptstyle\ \pm.00}$& $.04{\scriptstyle\ \pm.01}$& $.04{\scriptstyle\ \pm.01}$& $.79{\scriptstyle\ \pm.01}$ \\
        & \gcheck & \gcheck & \rcross & {\color{green!55!black}$\mathbf{\%}$} & $\mathbf{.18}{\scriptstyle\ \pm.00}$& $\mathbf{.07}{\scriptstyle\ \pm.01}$& $.07{\scriptstyle\ \pm.01}$& $.82{\scriptstyle\ \pm.01}$ & $\mathbf{.18}{\scriptstyle\ \pm.00}$& $\mathbf{.03}{\scriptstyle\ \pm.01}$& $\mathbf{.03}{\scriptstyle\ \pm.01}$& $.80{\scriptstyle\ \pm.01}$ \\
        & \gcheck & \gcheck & \gcheck & \rcross & $.19{\scriptstyle\ \pm.01}$& $\mathbf{.07}{\scriptstyle\ \pm.01}$& $.07{\scriptstyle\ \pm.01}$& $.80{\scriptstyle\ \pm.01}$  & $.21{\scriptstyle\ \pm.00}$& $.04{\scriptstyle\ \pm.01}$& $\mathbf{.03}{\scriptstyle\ \pm.01}$& $.74{\scriptstyle\ \pm.01}$ \\
        & \gcheck & \gcheck & \gcheck & {\color{green!55!black}\textbf{Qual.}} & $.19{\scriptstyle\ \pm.00}$& $.08{\scriptstyle\ \pm.01}$& $.08{\scriptstyle\ \pm.01}$& $.80{\scriptstyle\ \pm.01}$ & $.22{\scriptstyle\ \pm.00}$& $\mathbf{.03}{\scriptstyle\ \pm.01}$& $\mathbf{.03}{\scriptstyle\ \pm.01}$& $.70{\scriptstyle\ \pm.01}$ \\
        & \gcheck & \gcheck & \gcheck & {\color{green!55!black}$\mathbf{\%}$} & $\mathbf{.18}{\scriptstyle\ \pm.00}$& $\mathbf{.07}{\scriptstyle\ \pm.01}$& $.07{\scriptstyle\ \pm.01}$& $.81{\scriptstyle\ \pm.01}$ & $.20{\scriptstyle\ \pm.00}$& $\mathbf{.03}{\scriptstyle\ \pm.01}$& $\mathbf{.03}{\scriptstyle\ \pm.00}$& $.75{\scriptstyle\ \pm.01}$ \\
        \midrule
        \multirow{7}{*}{\rotatebox{90}{GPT-3.5 (black-box)}} 
         & \gcheck & \rcross & \rcross & \rcross & $\mathbf{.12}{\scriptstyle\ \pm.01}$& $.05{\scriptstyle\ \pm.01}$& $.05{\scriptstyle\ \pm.01}$& $.71{\scriptstyle\ \pm.03}$ & $.21{\scriptstyle\ \pm.00}$& $.03{\scriptstyle\ \pm.01}$& $.03{\scriptstyle\ \pm.01}$& $.72{\scriptstyle\ \pm.01}$ \\
        & \gcheck & \gcheck & \rcross & \rcross & $\mathbf{.12}{\scriptstyle\ \pm.01}$& $.06{\scriptstyle\ \pm.01}$& $.06{\scriptstyle\ \pm.01}$& $.72{\scriptstyle\ \pm.02}$ &  $\underline{\mathbf{.18}}{\scriptstyle\ \pm.01}$& $.04{\scriptstyle\ \pm.02}$& $.04{\scriptstyle\ \pm.02}$& $\underline{\mathbf{.82}}{\scriptstyle\ \pm.02}$ \\
        & \gcheck & \gcheck & \rcross & {\color{green!55!black}\textbf{Qual.}} &  $\mathbf{.12}{\scriptstyle\ \pm.01}$& $\mathbf{.03}{\scriptstyle\ \pm.01}$& $\mathbf{.03}{\scriptstyle\ \pm.01}$& $.72{\scriptstyle\ \pm.03}$ & $\mathbf{.18}{\scriptstyle\ \pm.01}$& $\mathbf{.02}{\scriptstyle\ \pm.01}$& $\mathbf{.02}{\scriptstyle\ \pm.00}$& $.80{\scriptstyle\ \pm.01}$ \\
        & \gcheck & \gcheck & \rcross & {\color{green!55!black}$\mathbf{\%}$} & $\mathbf{.12}{\scriptstyle\ \pm.01}$& $\underline{\mathbf{.03}}{\scriptstyle\ \pm.01}$& $\underline{\mathbf{.03}}{\scriptstyle\ \pm.01}$& $.72{\scriptstyle\ \pm.02}$ & $\mathbf{.18}{\scriptstyle\ \pm.00}$& $.04{\scriptstyle\ \pm.01}$& $.03{\scriptstyle\ \pm.00}$& $.80{\scriptstyle\ \pm.01}$ \\ 
        & \gcheck & \gcheck & \gcheck & \rcross & $\mathbf{.12}{\scriptstyle\ \pm.01}$& $.06{\scriptstyle\ \pm.01}$& $.06{\scriptstyle\ \pm.01}$& $.72{\scriptstyle\ \pm.02}$ & $.21{\scriptstyle\ \pm.00}$& $.03{\scriptstyle\ \pm.01}$& $.03{\scriptstyle\ \pm.01}$& $.72{\scriptstyle\ \pm.01}$ \\
        & \gcheck & \gcheck & \gcheck & {\color{green!55!black}\textbf{Qual.}} & $\mathbf{.12}{\scriptstyle\ \pm.01}$& $.04{\scriptstyle\ \pm.01}$& $.04{\scriptstyle\ \pm.01}$& $\underline{\mathbf{.73}}{\scriptstyle\ \pm.02}$ & $.21{\scriptstyle\ \pm.00}$& $.04{\scriptstyle\ \pm.01}$& $.04{\scriptstyle\ \pm.01}$& $.72{\scriptstyle\ \pm.01}$ \\
        & \gcheck & \gcheck & \gcheck & {\color{green!55!black}$\mathbf{\%}$} & $\mathbf{.12}{\scriptstyle\ \pm.01}$& $.04{\scriptstyle\ \pm.01}$& $.04{\scriptstyle\ \pm.01}$& $.64{\scriptstyle\ \pm.02}$ & $.21{\scriptstyle\ \pm.00}$& $\mathbf{.02}{\scriptstyle\ \pm.01}$& $\mathbf{.02}{\scriptstyle\ \pm.00}$& $.72{\scriptstyle\ \pm.01}$  \\
        \bottomrule
    \end{tabular}%
    }
    \caption{Calibration results for Vicuna v1.5 and GPT-3.5 on TriviaQA and CoQA using the auxiliary (clustering) method. We bold the best results per dataset, method and model.
    }\label{tab:calibration-results-features}
\end{table*}

\paragraph{Ablation results.} We show the results of the different ablations in \cref{tab:calibration-results-features}.
For both LLMs, we can see that the auxiliary model is able to achieve already decent scores by using the question as input alone. This suggest that the calibrator is learning about the general difficulty of questions for the LLM.
However, both cases also show that including more information---the LLM's answer, CoT reasoning or verbalized uncertainties---can help to improve the auxiliaries model calibration and misprediction AUROC even further, even though the effect remains somewhat inconsistent across models.

\subsection{Environmental Impact}\label{app:enviromental-impact}

All experiments are run on a single V100 NVIDIA GPU. 
Using \texttt{codecarbon} \citep{codecarbon, lacoste2019quantifying, lottick2019nergy}, we estimate finetuning the auxiliary calibrator on it to amount to 0.05 kgCO$_2$eq of emission with an estimated carbon efficiency of 0.46 kgCO$_2$eq / kWH.
Therefore, we estimate total emissions of around 1 kgCO$_2$eq to replicate all the experiments in our work.

\end{document}